\documentclass{article} 
\usepackage{iclr2027_conference,times}

\usepackage{amsmath,amsfonts,bm}

\def\eqref#1{equation~\ref{#1}}

\def\1{\bm{1}}

\DeclareMathAlphabet{\mathsfit}{\encodingdefault}{\sfdefault}{m}{sl}
\SetMathAlphabet{\mathsfit}{bold}{\encodingdefault}{\sfdefault}{bx}{n}

\usepackage{hyperref}
\usepackage{url}

\usepackage{amsmath,amssymb}
\usepackage{booktabs}
\usepackage{graphicx}
\usepackage{xcolor}
\usepackage{multirow}
\usepackage{tabularx}
\usepackage{subcaption}
\usepackage{enumitem}
\usepackage{longtable}
\usepackage{cleveref}

\usepackage[acronym]{glossaries}
\glsdisablehyper
\newacronym{dl}{DL}{Deep Learning}
\newacronym{nlp}{NLP}{Natural Language Processing}
\newacronym{qkv}{QKV}{Query, Key, Value}
\newacronym{cls}{[CLS]}{Classify token}
\newacronym{fm}{FM}{Foundation Model}
\newacronym{sota}{SOTA}{State-of-the-art}
\newacronym{e2e}{E2E}{End-to-End}

\newacronym{nn}{NN}{Neural Network}
\newacronym{icl}{ICL}{In-context-learning}
\newacronym{gbdt}{GBDT}{Gradient-Boosted Decision Tree}
\newacronym{lgbm}{LightGBM}{Light Gradient-Boosting Machine}
\newacronym{tfm}{TFM}{Tabular Foundation Model}
\newacronym{kl}{KL}{Kullback-Leibler}

\newacronym{auroc}{AUROC}{Area Under the Receiver Operating Characteristic Curve}

\newacronym{lt}{LT}{Linear Transformation}
\newacronym{id}{Id}{Identity}

\title{Tabular Foundation Models Are Effectively Shallow}

\author{Irene Cannistraci, Julia E. Vogt \\
Department of Computer Science\\
ETH Zurich\\
Switzerland\\
\texttt{irene.cannistraci@inf.ethz.ch} \\
}

\iclrfinalcopy 
\begin{document}
\maketitle

\begin{abstract}
\looseness=-1In-context-learning \glspl{tfm} are uniquely suited to scenarios with limited resources or privacy constraints, as they solve new tasks without additional training. However, their architecture consists of multiple transformer blocks, resulting in high computational and memory footprints. Meanwhile, traditional alternatives like \glspl{gbdt} remain highly competitive and run efficiently on CPUs, suggesting that tabular tasks may not inherently require deep architectures. This contrast raises a fundamental question: \emph{can \glspl{tfm} be simplified by removing architectural redundancies without sacrificing performance?} To answer this, we introduce a closed-form, label-free affine translator that replaces a contiguous window of up to 96\% of the blocks, leaving the rest unchanged. Evaluating eight frontier \glspl{tfm} on 171 classification datasets, we reveal that tabular transformers are vastly overparameterized: replacing half of a model's blocks costs at most $0.002$ median AUROC across all architectures, with the half-depth TabFM still outperforming every other full-depth \gls{tfm} and tuned \glspl{gbdt}. Furthermore, different model families are overparameterized in different ways: TabPFN-3 relies on its encoders, maintaining performance regardless of which single block is retained. Other single-axis models spread computation thinly, needing only their first block, while dual-axis models concentrate it in one critical block. Ultimately, removing depth saves up to 1.5B parameters, speeds up GPU inference by up to $6\times$ at a median cost of $0.03$ AUROC, while a single-block TabPFN-3 serves cached queries $2.4\times$ faster using $6\times$ less memory at negligible cost.
\end{abstract}

\section{Introduction} \label{sec:intro}
\begin{figure}[h]
    \centering
    \includegraphics[width=\textwidth]{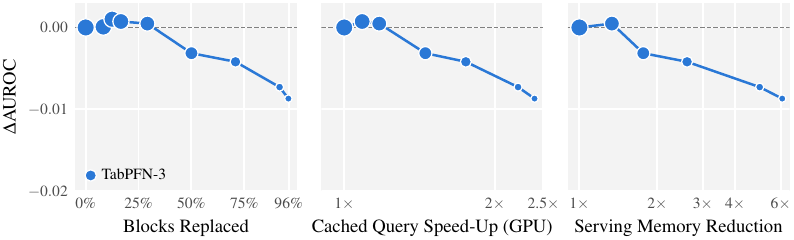} 
    \caption{\textbf{TabPFN-3 at different levels of compression.} Change in the classification AUROC using MIMIC-III vs. (left) \% of replaced blocks, (middle) cached-query GPU speed-up, and (right) serving-memory reduction. Keeping only one transformer block (96\% of blocks replaced) costs $-0.009$ AUROC for a $2.4\times$ speed-up and $6.1\times$ less memory.}
    \label{fig:teaser}
\end{figure}
\gls{icl} \glspl{tfm} offer a powerful approach for tabular prediction: they solve new tasks with no additional training, handle heterogeneous data, and require no preprocessing or hyperparameter tuning. Recent models such as TabPFN~\citep{hollmann2025accurate,grinsztajn2026tabpfn}, TabICL~\citep{qu2025tabicl}, and TabDPT~\citep{ma2025tabdpt} achieve strong performance across diverse benchmarks. However, this zero-shot convenience comes at a high computational cost. Current \glspl{tfm} stack 12 to 24 identical transformer blocks on top of their encoders. This depth imposes heavy memory footprints and high inference latency, limiting deployment on resource-constrained hardware such as local clinical servers. 
Meanwhile, traditional alternatives like \glspl{gbdt} remain highly competitive on these same tasks while running efficiently on CPUs~\citep{grinsztajn2022tree,mcelfresh2023neural}. The success of these shallower, tree-based models suggests that typical tabular tasks do not inherently require deep hierarchical composition. If that is true, much of the deep architecture inside current \glspl{tfm} may be computationally redundant. This contrast raises a fundamental question: \emph{can we retain the zero-shot convenience of \glspl{tfm} while approaching the efficiency of simpler models by removing architectural redundancy?}

To answer this, we must first understand how these models utilize their depth. While observational metrics such as pairwise representation similarity and attention profiling reveal signs of architectural redundancy, the correspondence between these signals and actual replaceability is inconsistent across models. We therefore locate redundancy by intervening on the architecture itself: we introduce a closed-form, label-free affine translator that bypasses a contiguous window of transformer blocks. If the output distribution remains stable, the skipped blocks were performing computations that a simple linear map can absorb; if the output collapses, they were performing necessary non-linear composition. We design this translator with a per-feature parameterization to support dual-axis attention, and select the optimal bypass window by minimizing the forward KL divergence on a small calibration split\footnote{code is available at \url{https://github.com/icannistraci/shallow-tfm}}. 

Evaluating 8 \glspl{tfm} from 6 model families across 171 classification datasets, we show that current tabular architectures are overparameterized. Replacing half the blocks across all models results in a minimal median drop of at most $0.002$ AUROC, with up to 97\% of datasets staying within one point of the full model's performance. Pushing compression to a single block maintains high performance for TabPFN-3, Orion-Bix, and TabICL, which stay within a median $0.007$ AUROC of their full-depth counterparts. The savings translate directly to hardware efficiency: halving the depth doubles the throughput of the transformer stack, and with a cached context, per-query GPU latency drops by $1.5\times$--$1.9\times$ at Half and $2.4\times$--$5.1\times$ with one block kept. We show that naively dropping blocks causes model collapse, and early exit performs worse than the label-free translator in most configurations. The translator can also be reused across similar datasets. For example, a half-depth TabICL translator fitted on MIMIC-III~\citep{johnson2016mimic} transfers to the eICU~\citep{pollard2018eicu} dataset with an AUROC penalty of just $0.004$ (against $0.003$ for a local fit), though transferability diminishes between datasets with different schemas (e.g., Heart and Diabetes, \Cref{sec:transfer}).

Finally, through a controlled pre-training study on nanoTabPFN \citep{pfefferle2025nanotabpfn}, we investigate whether compression matches natively shallow training. We observe that a 12-layer model compressed to $k$ layers matches the performance of a model trained from scratch at depth $k$ (within $0.011$ AUROC). Furthermore, a natively trained 6-layer model matches the AUROC of the full 12-layer baseline, indicating that depth is over-provisioned in this specific setting. Whether this structural redundancy scales to significantly deeper architectures remains an open question for future work. For existing models, our proposed approach offers a practical method to achieve the inference efficiency of a shallow architecture without the computational cost of retraining from scratch.

\section{Related work}
\label{sec:related}

\paragraph{Tabular Transformers.} The development of tabular transformers began with task-specific architectures. Early adaptations like FT-Transformer~\citep{gorishniy2021revisiting} and SAINT~\citep{somepalli2021saint} apply standard attention mechanisms to supervised learning, while TabNet~\citep{arik2021tabnet} uses sequential sparse routing for step-wise feature selection. Since then, the field has moved toward general-purpose foundation models, with a strong focus on in-context learning \glspl{tfm}. TabPFNv2~\citep{hollmann2025accurate} and TabPFN-3~\citep{grinsztajn2026tabpfn} scale prior-data fitted networks to handle heterogeneous columns and mixed data types. TabICL~\citep{qu2025tabicl, qu2026tabiclv2} processes tabular structures via set-transformers. TabDPT~\citep{ma2025tabdpt} pre-trains on real data, while TabFM~\citep{google2026tabfm} is pre-trained on synthetic data, and Mitra~\citep{zhang2026mitra} mixes diverse synthetic priors. Finally, the Orion family~\citep{bouadi2025orion, bouadi2026orion} and LimiX~\citep{zhang2025limix} vary the attention pattern of the same in-context recipe (multi-scale sparse, bi-axial and dual-axis cell attention).

\paragraph{Transformer Redundancy.} \looseness=-1Over-parameterization is a well-documented trait of vision and language transformers, where dropping layers, attention heads, or tokens often yields minimal performance degradation~\citep{gromov2025unreasonable, men2024shortgpt, michel2019sixteen, bolya2023tome, venkataramanan2024skip}. In vision, \citet{jacobs2026block} replace deep transformers with a small number of recurrent blocks, while \citet{bolya2023tome} explicitly exploit token-level redundancy. More closely related to our approach, \citet{shopkhoev2025replaceme} and \citet{cannistraci2024toast} show that training-free linear maps can accurately bridge skipped block representations. Following the recent emergence of \gls{icl} \glspl{tfm}, several concurrent works have begun exploring depth and context redundancy in the tabular domain. However, concurrent approaches fundamentally differ from ours by requiring either additional pre-training or dataset-specific gradient updates. For instance, \citet{balef2026onelayer} pre-train a single-layer model from scratch to loop at inference, while \citet{zabergja2026taco} pre-train a predictor-compressor pair to condense context rows into latent vectors. Among post-training approaches, TACTICL \citep{koshil2026tacticl} replaces transformer layers with lightweight MLP adapters trained via gradient descent on the downstream task. Similarly, \citep{tanna2026distill} relies on labeled data to train task-specific student models, while early-exit methods \citep{kueken2025earlyexit} pre-train new layer-wise decoders on prior data. In contrast, our approach uses a single, closed-form affine map, requiring no non-linear adapters, gradient updates, or labeled data

\section{Method} \label{sec:method}

Let $\mathbf{H}_l \in \mathbb{R}^{N \times d}$ denote the latent representations at depth $l \in \{0, \dots, L\}$, where $\mathbf{H}_0$ represents the output of the input embedding layer and $\mathbf{H}_L$ is the output of the final transformer block. For a contiguous window of blocks $0 \leq s < e \leq L$, we bypass blocks $s$ through $e{-}1$ entirely. We replace this bypassed window with an affine map $(\mathbf{W}^*, \mathbf{b}^*)$, fitted via ridge regression on an unlabeled calibration set:
\begin{equation}
    (\mathbf{W}^*, \mathbf{b}^*) = \arg\min_{\mathbf{W},\mathbf{b}} \|\mathbf{H}_e - (\mathbf{H}_s \mathbf{W} + \mathbf{b})\|_F^2 + \lambda\,\|\mathbf{W}\|_F^2.
    \label{eq:ridge}
\end{equation}

The hidden states $\mathbf{H}_s$ and $\mathbf{H}_e$ are extracted using a single forward pass over the in-context support set and the unlabeled calibration queries. To bound the solve time independently of context size, we fit the map on a fixed random subset of $N = \min(500, n_\text{support} + n_\text{cal})$ of these sequence positions. The regularization parameter $\lambda \in \{0, 10^{-3}, 10^{-2}, 10^{-1}\}$ is scaled relative to the mean diagonal of the Gram matrix. We evaluate the fitting residual on a 25\% held-out validation split of these calibration rows to select $\lambda^*$, then refit the map on the full subset. 

At inference, this closed-form map acts as a linear module projecting $\mathbf{H}_s$ directly to $\mathbf{H}_e$, while the remaining blocks $e, \dots, L{-}1$ and the original classification head operate entirely unchanged. The closed-form fit adds negligible overhead to context encoding. Additionally, refining the map by gradient descent does not improve it (\Cref{app:refinement}).

\paragraph{Per-Architecture Parameterization.} Single-axis self-attention architectures such as TabFM and TabPFN-3 operate on sequence tensors where items are processed independently, requiring a single shared transformation matrix $(\mathbf{W}, \mathbf{b})$. Conversely, dual-axis models such as TabPFNv2 and LimiX attend across both rows and features, producing hidden tensors of shape $(\cdot, n_\text{features}, d)$. Applying a single shared matrix across this tensor would mix feature representations and disrupt axis alignment. Thus, for dual-axis architectures, we fit a distinct affine map $(\mathbf{W}_f, \mathbf{b}_f)$ for each feature index $f \in \{1, \dots, n_\text{features}\}$ (additional details in \Cref{app:translator-family}). This expands the total translator parameterization to $n_\text{features}(d^2{+}d)$, which scales dynamically with the dataset's feature dimensionality. However, in some cases (e.g., TabPFNv2), these parameters can be largely absorbed.

\paragraph{Absorbing the Map.} Because the map is affine, it can be absorbed into an adjacent affine layer. However, since transformer blocks contain non-linear operations, this operation is restricted to the network boundaries. If the window ends at the final block and the classification head applies a linear projection $\mathbf{x} \mapsto \mathbf{x}\mathbf{A}^\top + \mathbf{c}$, the map folds exactly as $\mathbf{H}_s(\mathbf{W}^*\mathbf{A}^\top) + (\mathbf{b}^*\mathbf{A}^\top + \mathbf{c})$. Symmetrically, if the window starts at block $0$, the map folds into the encoder's final linear layer. In these cases, the compressed model strictly reduces to a standard $k$-block architecture with no additional parameters. Additional details are provided in \Cref{app:folding}.

\paragraph{Label-Free Window Selection.} To determine which block window to bypass, we evaluate candidate window bounds $(s, e)$ on the calibration split $X_\text{cal}$. Because our bypass retains the original pre-trained classification head, candidate models produce directly comparable probability distributions. Inspired by the KL-guided layer selection of \citet{li2026distilling}, for a target width $k$, we select the optimal window $(s^*, e^*)$ by minimizing the forward KL divergence between the uncompressed model's output distribution $p_\text{full}$ and the compressed model's distribution $p_{(s,e)}$:
\begin{equation}
    (s^*, e^*) = \arg\min_{(s,e):\, e-s=k} \, \mathbb{E}_{x \in X_\text{cal}} \left[ \mathrm{KL}\left( p_\text{full}(y \mid x) \,||\, p_{(s,e)}(y \mid x) \right) \right].
    \label{eq:kl_selection}
\end{equation}
Minimizing output KL divergence rather than the latent residual (e.g., $\|\mathbf{H}_e - \mathbf{\hat{H}}_e\|_F^2$) is advantageous for two reasons. First, latent distances weigh all error directions equally, whereas the KL divergence weighs each error by its effect on the final predictions, after the remaining blocks and the classification head have amplified or suppressed it. Second, hidden-state scales differ across depths, so latent residuals of windows ending at different blocks are not comparable; the KL divergence scores every candidate in the same output probability space, across window positions and widths.

Window selection is executed as a one-time pre-computation during context encoding. Candidate windows share a single hooked forward pass over $X_\text{cal}$ to cache intermediate states. For each candidate $(s,e)$, fitting the affine map and scoring the resulting distribution requires only a closed-form matrix solve and a fast pass through the remaining $L-e$ layers. Once selected, the affine map is frozen for all subsequent queries. Single-axis models permit a static default selection, whereas dual-axis ones exhibit dataset-specific layer sensitivities that benefit from dynamic selection (\Cref{tab:fixed_window}). The same criterion can also choose how many blocks to replace. Given a divergence tolerance $\epsilon > 0$, we select the widest window whose calibration drift stays within $\epsilon$:
\begin{equation}
    \max_{s, e} \, (e - s) \quad \text{s.t.} \quad \mathbb{E}_{x \in X_\text{cal}} \left[ \mathrm{KL}\left( p_\text{full} \,||\, p_{(s,e)} \right) \right] \le \epsilon.
    \label{eq:tolerance}
\end{equation}
At $\epsilon = 0.01$, this rule removes 52--70\% of the blocks of the single-axis models at a mean cost of at most $0.004$ AUROC (\Cref{app:kl_frontier}).

\section{Locating Computational Redundancy} \label{sec:analysis}

Before evaluating the performance of compressed models, we investigate how computation is distributed across the transformer blocks. We first examine representation and attention dynamics, then we perform a per-block intervention to accurately locate redundant blocks to replace.

\paragraph{Experimental Setup.} We evaluate 8 pre-trained \glspl{tfm} from 6 model families (TabPFN, TabICL, TabDPT, TabFM, Orion and LimiX) with frozen weights across 171 classification datasets: 163 from four deduplicated benchmark suites (OpenML-CC18~\citep{bischl2017openml}, TabZilla~\citep{mcelfresh2023neural}, TabArena~\citep{erickson2025tabarena}, Grinsztajn~\citep{grinsztajn2022tree}) and eight clinical datasets including MIMIC-III~\citep{johnson2016mimic} and eICU-CRD~\citep{pollard2018eicu}. Each dataset is split into train, calibration, and test, and run at 3 seeds. For deterministic extraction, where possible, prediction-time ensembling is reduced to a single member, and LimiX-16M's input preprocessing is fitted once on the support rows and then frozen (\Cref{app:translator-family}). Additional details, including datasets and models details, are in \Cref{app:setup}, while additional results are in \Cref{app:additional_experiments}.

\begin{figure}[h]
    \centering
    \includegraphics[width=\columnwidth]{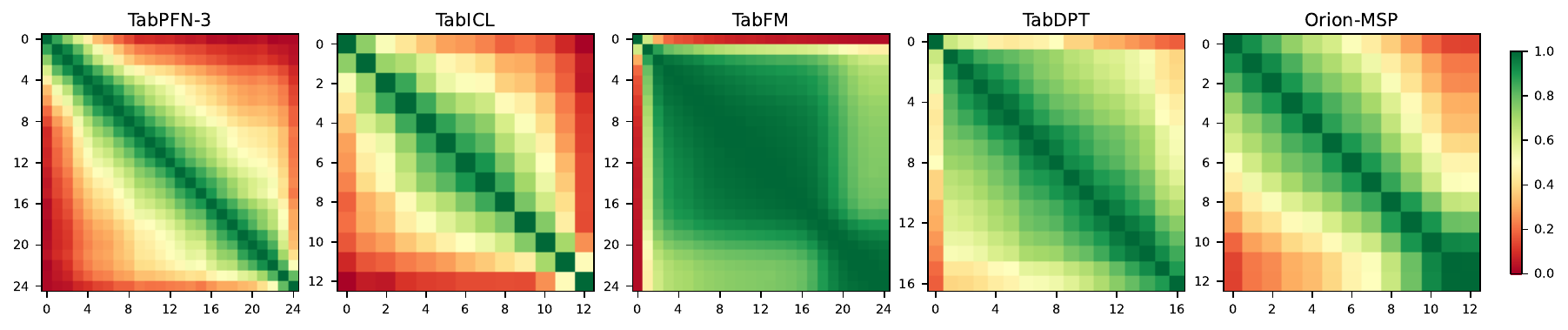}
    \caption{\textbf{Pairwise Cosine Similarity.} Layerwise cosine similarity between all block representations using MIMIC-III across different \glspl{tfm}. Except for TabFM, models show architecture-specific structure that does not reliably predict which blocks are replaceable.}
    \label{fig:layer_cosine}
\end{figure}

\paragraph{Representation Similarity.} \Cref{fig:layer_cosine} shows the full pairwise cosine similarity between every pair of block representations on a single forward pass using MIMIC-III. The resulting matrices vary drastically across architectures. For example, from block~3 on, consecutive TabFM representations have cosine similarity ${}\geq 0.99$, consistent with the concurrent findings of~\citet{balef2026onelayer} that other \glspl{tfm} representations stabilize early. In contrast, TabDPT shows broad similarity in its middle blocks and Orion-MSP in its last ones, while TabPFN-3 exhibits a smooth transition with no sharp boundaries. Finally, TabICL's final block is the most dissimilar from every other block, while results in \Cref{fig:sensitivity_full} show that it is fully replaceable by the affine translator. Representation similarity therefore does not accurately predict replaceability.

\begin{figure}[h]
    \centering
    \includegraphics[width=.85\columnwidth]{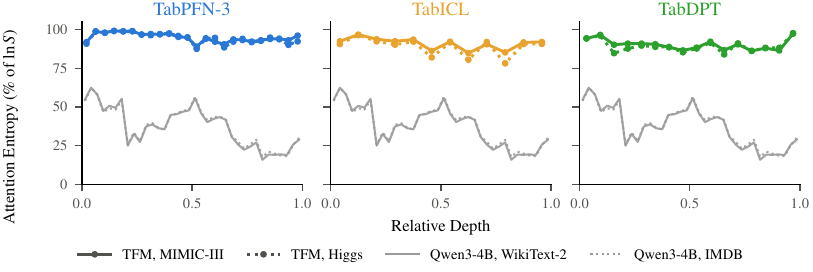}
    \caption{\textbf{Per-block Attention Entropy.} Normalized attention entropy (percentage of maximum $\ln S$) for three representative \glspl{tfm} (on MIMIC-III and Higgs) vs.\ Qwen3-4B. While tabular models attend near-uniformly to all tokens, Qwen3-4B displays more peaked attention.}
    \label{fig:nlp_comparison}
\end{figure}

\paragraph{Attention Patterns.} If the depth of a tabular \gls{tfm} is redundant, one might expect later blocks to simply stop specializing and repeatedly retrieve the same context rows. However, attention statistics on two diverse datasets, MIMIC-III and Higgs, show a different behaviour: the five most-attended keys per query overlap by a Jaccard index of only $0.10$--$0.19$ between consecutive blocks (vs.\ $0.005$ randomly). Additionally, TabPFN-3's first block shares almost no keys with later ones (\Cref{app:attention_entropy_def}). Nor does attention specialize with depth as in language models. \Glspl{tfm} attend almost uniformly over their context at every block (averaging 81--95\% of maximum entropy, and $\ge$70\% at every block except TabFM's read-out block), whereas Qwen3-4B's attention is sharply peaked and highly variable (\Cref{fig:nlp_comparison} and \Cref{app:nlp_comparison}). Ultimately, although each block retrieves a distinct set of rows, the entropy profile remains persistently flat. Consequently, we must look beyond attention mechanisms or embedding similarities to understand when the network becomes redundant. 

\begin{figure}[h]
    \centering
    \includegraphics[width=\textwidth]{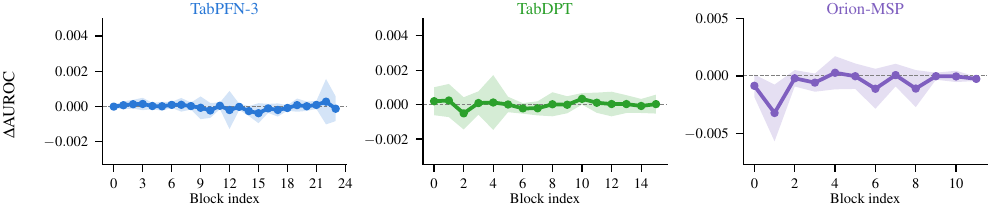}
    \caption{\textbf{Per-block Sensitivity.} $\Delta$AUROC when a single block is replaced by the translator. Mean across 8 clinical datasets and 3 seeds. Full results in \Cref{fig:sensitivity_full}.}
    \label{fig:sensitivity}
\end{figure}

\paragraph{Per-Block Sensitivity.} \label{sec:blocks-analysis} To locate the redundant blocks directly, we use the affine translator itself as diagnostic tool. Specifically, we intervene by replacing individual blocks with the translator and measuring the resulting change in AUROC (\Cref{fig:sensitivity}). Notably, the surviving blocks behave just as they do in the full model (attention entropy shifts by ${\leq}\,2\%$; \Cref{app:attention}). For single-axis models, no single block is critical and every replacement stays within $0.007$ AUROC of the full model. In contrast, dual-axis models each have exactly one critical block (i.e., block 0 for TabPFNv2 and block 3 for LimiX-16M), while every other block stays within $0.043$ and $0.009$, respectively. In short, evaluated one block at a time, almost the entire stack is affine-replaceable and the non-affine computation that remains sits in a single block of the dual-axis models. More results in \Cref{fig:sensitivity_full}.

\section{Evaluation of Compressed Models} \label{sec:main-results}

Building on \Cref{sec:analysis}, we now evaluate the predictive performance of our compressed models. Although we analyze the impact of replacing contiguous block windows of varying widths ranging from 0\% to 96\% of the network (\Cref{fig:teaser,fig:compression_per_model}), for clarity we focus our discussion on two primary configurations: (i) \emph{Half}, a safe, deployable setting that replaces 50\% of the blocks with the affine translator, and (ii) \emph{One-block}, an extreme setting that retains only a single block alongside the translator. Additional results are in \Cref{tab:compact,tab:foundation}.

\subsection{Classification} \label{sec:classification}

\begin{figure}[h]
    \centering
    \includegraphics[width=\columnwidth]{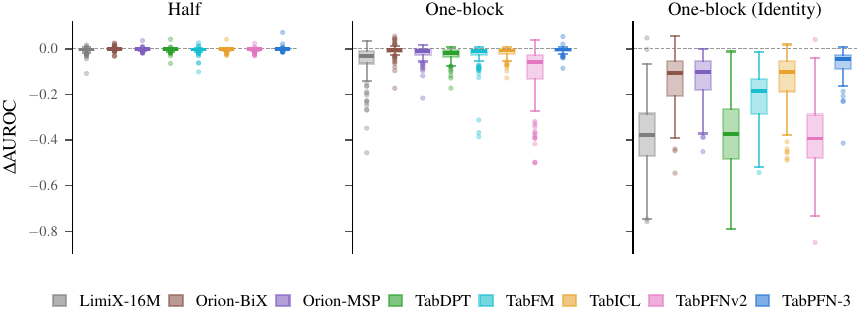}
    \caption{\textbf{Classification Performance at Varying Compression Levels.} Per-dataset $\Delta$AUROC (compressed $-$ full) over 163 datasets at half-depth (left), reduced to a single block using our translator (center), and a single-block baseline using an identity mapping (right). Additional results in \Cref{tab:compact,tab:foundation}.}
    \label{fig:cdf}
\end{figure}

\Cref{fig:cdf} shows the change in AUROC (compressed $-$ full) on each of the 163 suite datasets, averaged over three seeds, for every model at Half, at One-block, and at One-block with the identity in place of the translator. In the left panel (\emph{Half}), replacing 50\% of the transformer stack with the single affine translator causes practically no performance loss. Across all architectures and dataset scales, the mean AUROC drops by a negligible $0.001$ to $0.005$, with median losses of at most $0.002$. The stability holds on the larger datasets too, and the ranking of the models, among themselves and against tuned \glspl{gbdt}, is unchanged (\Cref{app:suite_ranks}). Discrete metrics reflect the same stability: accuracy and balanced accuracy drop by at most $0.010$ and $0.018$ across all tasks (\Cref{app:decision_metrics}).

With a single block kept (\emph{center}), the models separate into three groups. TabPFN-3 and Orion-Bix remain nearly lossless, while  TabICL, Orion-MSP, TabFM, and TabDPT suffer negligible mean losses ($0.015$ to $0.026$) with some visible outliers, and dual-axis models degrade more, particularly on large datasets and long contexts (\Cref{app:context}). TabPFNv2 suffers the most (mean drop $0.101$), and LimiX-16M degrades more mildly (mean drop $0.052$).

Additionally, this extreme compression isolates exactly where different architectures perform their core computations (\Cref{tab:kept_block}). Single-axis models strongly prefer retaining their first block, relying on it for the initial non-linear attention pass over the context rows. Conversely, dual-axis models must retain their last block to ensure the final representations match the exact distribution expected by the classification head. With a single contiguous window, the kept block is either the first or the last. For TabPFNv2, keeping its critical block~0 costs $0.281$ AUROC, while replacing it and keeping the last block costs only $0.071$ AUROC. LimiX-16M's critical block~3 is neither first nor last. The map must therefore absorb the critical block's computation, which is why the dual-axis models lose the most at \emph{One-block}; keeping the critical block between two maps would require two windows, which we leave as feature work. TabPFN-3 remains indifferent, as its core computation is already localized in its encoders. Finally, the failures of \emph{One-block} indicate when depth is needed. The cost grows with dataset size (Spearman $\rho=-0.22$; \Cref{app:context}), and the largest loss averaged over the six single-axis models is on \texttt{monks-2} ($-0.128$), whose label is a logical rule over its features. 

\paragraph{Ablations.} One could argue that the skipped blocks are simply inactive, making our translator unnecessary. To test this, we evaluate dropping the same blocks without applying the affine map, effectively treating the skipped section as an identity function (\Cref{fig:cdf}, \emph{right}). Under this baseline, performance collapses for most models down to chance level (\Cref{tab:compact}). The skipped blocks are therefore not idle, but the closed-form map recovers their computation.

Because an affine approximation suffices, we also evaluate a standard early-exit baseline: fitting a supervised logistic regression classifier directly to the representations right before the skipped window, thereby discarding the remaining blocks and the original head. Despite having access to the support labels, this shortcut underperforms in most configurations. It loses $\approx 0.02$ AUROC at every depth, whereas our label-free translator matches or outperforms it in 81\% of the 2{,}664 tested configurations (\Cref{tab:early_exit_summary}). Finally, repeatedly looping the retained block also fails to substitute the affine map (\Cref{tab:kept_block}). Mapping intermediate representations into the space the remaining blocks and the frozen head expect therefore beats training a new head from scratch, even though the new head sees labels.

\subsection{Regression} \label{sec:regression}

Since \Cref{sec:classification} shows that up to 96\% of a tabular transformer's depth can be safely removed, we now investigate whether this extreme redundancy is a general property of tabular models or specific to classification. We evaluate the regressor variants of our architectures across 41 continuous datasets. For these experiments, we select the skip window using calibration $R^2$, as regressors lack the output distributions necessary for the KL divergence criterion. Additional results in \Cref{tab:regression}.

\begin{figure}[h]
    \centering
    \includegraphics[width=\columnwidth]{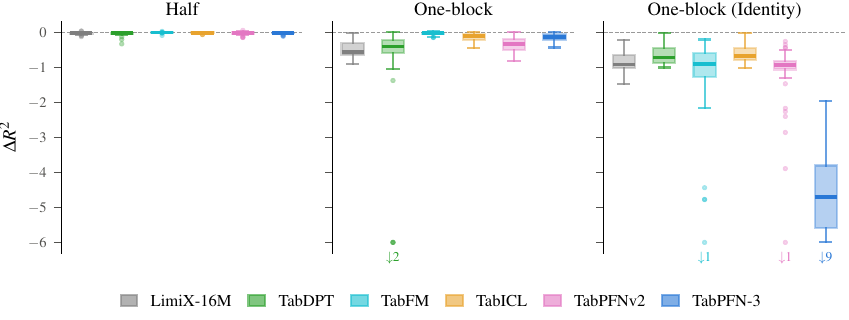}
    \caption{\textbf{Regression Performance at Varying Compression Levels.} Per-dataset $\Delta R^2$ (compressed $-$ full) over 41 regression datasets at half-depth (left), reduced to a single block using our translator (center), and a single-block baseline using an identity mapping (right).}
    \label{fig:regression}
\end{figure}

Regression shows the same redundancy at \emph{Half} (\Cref{fig:regression}): replacing half of the blocks costs only $0.001$ to $0.026$ in median $R^2$, whereas dropping them without the map costs far more (\emph{right}). With one block kept, however, most regressors degrade sharply (e.g., TabPFN-3 loses $0.132$ $R^2$), with TabFM the exception ($-0.022$ despite going from 24 blocks to 1). Two effects contribute. The predictions shrink toward the mean (standard deviation $0.48$--$0.72$ of the full model's; $0.22$ for LimiX-16M), which an affine recalibration on the calibration split partly repairs (\Cref{tab:regression_scale}). The samples are also ranked less accurately (Spearman $\rho$ from $0.93$ to $0.86$ for TabICL and from $0.91$ to $0.57$ for TabDPT). The late blocks of a regressor therefore refine the ordering of its outputs, which regression needs but classification does not, as long as the decision boundary is preserved. This is why classifiers survive a single block while most regressors need about half of their depth.

\section{Efficiency} \label{sec:params}
\looseness=-1Finally, we analyze how this compression affects computational cost. \Cref{tab:hardware} summarizes the hardware impact of compression on a single RTX 5090. Because the transformer stack accounts for 93\%--98\% of total weights in most \glspl{tfm} (except Orion-MSP, at 57\%), dropping blocks removes most of the weights. For single-axis models, the affine translator adds negligible overhead (e.g., 262k parameters replacing 23.6M in TabICL), allowing \emph{One-block} compression to eliminate 91\%--96\% of stack weights. At this extreme depth, TabFM sheds roughly 1.5B parameters. Dual-axis models present an exception: because their affine map scales with the number of feature tokens, the translator overhead grows on wider datasets, moderating the net parameter reductions (yielding 70\%--71\% stack savings at \emph{One-block} on MIMIC-III). This map overhead can often be absorbed by adjacent components (\Cref{sec:method}). For example, TabPFNv2 folds its \emph{One-block} map exactly into the encoder, shrinking it from 1.45M to 43k parameters (boosting stack savings from 71\% to 91\%), while TabDPT folds its 0.59M-parameter map into the classification head. Other models require normalization at these boundaries, retaining their already small maps. For LimiX-16M, whose encoder ends in a LayerNorm, the map cannot be absorbed and grows by 37k parameters per feature token, so on very wide datasets \emph{One-block} LimiX-16M measures redundancy rather than saving parameters.

\begin{table}[h]
    \centering
    \small
    \caption{\textbf{Hardware Impact.} MIMIC-III, single RTX 5090. Blocks params: parameters of the transformer blocks kept, translator included. $\Delta$FLOPs: forward-pass FLOPs removed. Re-encode: batch of 512 queries, context re-encoded on every call. Cached: one query against a cached context (three models); memory = weights + cache.}
    \label{tab:hardware}
    \small
    \setlength{\tabcolsep}{4pt}
    \resizebox{\textwidth}{!}{%
        \begin{tabular}{l ccc cc cc cc cc}
            \toprule
                      & \multicolumn{3}{c}{Blocks params (M)} & \multicolumn{2}{c}{$\Delta$FLOPs} & \multicolumn{2}{c}{Re-encode speed-up} & \multicolumn{2}{c}{Cached query speed-up} & \multicolumn{2}{c}{Cached memory reduction} \\
            \cmidrule(lr){2-4} \cmidrule(lr){5-6} \cmidrule(lr){7-8} \cmidrule(lr){9-10} \cmidrule(lr){11-12}
            Model     & Full & Half & One-block & Half & One-block & Half & One-block & Half & One-block & Half & One-block \\
            \midrule
            TabPFNv2 & 7.1 & 5.0 & 2.0 & $-$50\% & $-$91\% & 1.7$\times$ & 4.6$\times$ & 1.9$\times$ & 5.1$\times$ & 1.9$\times$ & 10.0$\times$ \\
            TabPFN-3 & 51.4 & 25.9 & 2.4 & $-$26\% & $-$51\% & 1.1$\times$ & 1.2$\times$ & 1.5$\times$ & 2.4$\times$ & 1.8$\times$ & 6.1$\times$ \\
            TabICL & 25.7 & 13.1 & 2.4 & $-$23\% & $-$42\% & 1.0$\times$ & 1.1$\times$ & 1.7$\times$ & 4.0$\times$ & 1.7$\times$ & 4.3$\times$ \\
            TabDPT & 75.6 & 38.4 & 5.3 & $-$49\% & $-$92\% & 1.5$\times$ & 3.3$\times$ & --- & --- & --- & --- \\
            TabFM & 1{,}611 & 810 & 71.3 & $-$31\% & $-$59\% & 1.2$\times$ & 1.4$\times$ & --- & --- & --- & --- \\
            Orion-MSP & 25.2 & 12.9 & 2.4 & $-$22\% & $-$37\% & 1.1$\times$ & 1.2$\times$ & --- & --- & --- & --- \\
            Orion-Bix & 25.2 & 12.9 & 2.4 & $-$21\% & $-$40\% & 1.2$\times$ & 1.3$\times$ & --- & --- & --- & --- \\
            LimiX-16M & 15.9 & 11.4 & 4.8 & $-$50\% & $-$91\% & 1.8$\times$ & 6.1$\times$ & --- & --- & --- & --- \\
            \bottomrule
        \end{tabular}}
\end{table}

While the compressed blocks run up to $23\times$ faster (\Cref{tab:stack_latency_gpu}), the end-to-end speed-up is limited by the time spent outside them, mostly in the encoders (Amdahl's law). At \emph{One-block}, models with lightweight encoders (TabPFNv2, TabDPT, LimiX-16M) accelerate by $3.3\times$--$6.1\times$, whereas heavy encoders (TabICL, TabFM) limit gains to $1.1\times$--$1.4\times$; for these models the main gain is memory, e.g.\ \emph{One-block} TabFM drops about 6\,GB of fp32 weights. Context caching removes this bottleneck, yielding query speed-ups of up to $5.1\times$. On CPUs, where the blocks dominate the run time, end-to-end speed-ups reach $12.4\times$ (\Cref{tab:hardware_cpu}). The same gains apply to regression, which uses the same backbones: at \emph{Half}, where regression is near-lossless, the lightweight models accelerate by $1.5\times$--$1.8\times$ on GPU, and at \emph{One-block} TabFM sheds 1.5B parameters for a median $\Delta R^2$ of only $-0.022$.

\section{Compression vs. Pre-Training Shallow Architectures}
If up to 96\% of a tabular transformer's layers can be safely approximated, a natural follow up question would be: why not simply pre-train a shallow model from scratch? To analyze this, we test whether compressing a model matches the performance of a natively shallow one using the nanoTabPFN reference implementation \citep{pfefferle2025nanotabpfn}\footnote{\texttt{github.com/automl/nanoTabPFN}}. We train the model across varying depths $L \in \{1,2,3,6,12\}$, each with an embedding dimension of 96 and 4 attention heads, for 5,000 steps across three random seeds (see \Cref{app:nano_setup} for configurations). We evaluate these models on four binary clinical datasets (Diabetes, Heart, Breast Cancer, SUPPORT2), strictly operating in the prior-fitting regime (100 support rows, top 5 F-score features, 1,000 test rows).  We compare these natively shallow models against the 12-layer model compressed to $k$ blocks using our per-feature translator. Although scaled down (1.4M parameters), this proxy model provides a controlled environment to study the architectural origins of redundancy.

\begin{figure}[h]
    \centering
    \includegraphics[width=.9\textwidth]{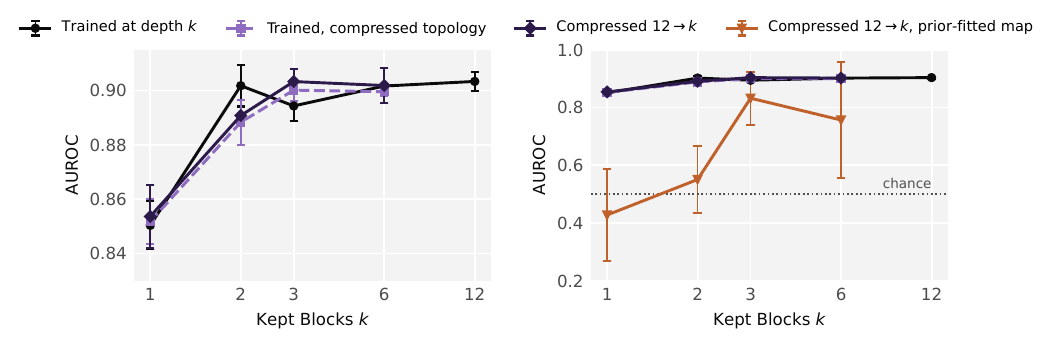}
    \caption{\textbf{Pre-training Redundancies.} nanoTabPFN test AUROC, averaged over four clinical datasets. Compressing 12 blocks to $k$ matches native training at depth $k$, but only when the translator is fitted on the target's (unlabelled) rows rather than on prior tasks (right). Exact values in \Cref{tab:train_shallow}.}
    \label{fig:train_shallow}
\end{figure}

Within this controlled experimental setting, results in \Cref{fig:train_shallow} suggest three properties of \glspl{tfm}. First, compressing a pre-trained model performs on par with pre-training a shallow one: AUROC matches within $0.011$ at every depth, including a single block. Second, this parity does not come from the translator's extra parameters: training the compressed architecture from scratch, i.e., $k$ blocks plus an identity-initialized $d \times d$ layer (``trained, compressed topology''), does not reliably beat the plain $k$-block model ($0.888$ vs.\ $0.902$ at $k=2$; $0.900$ vs.\ $0.894$ at $k=3$). It does, however, require fitting the map to the target: fitting the map on 20 unseen prior tasks instead of the target's rows drops AUROC to $0.43$--$0.83$ (\Cref{fig:train_shallow}, right), consistent with the transfer study (\Cref{sec:transfer}). Third, depth is over-provisioned: a natively trained 6-block model matches the 12-block one, and deeper models are harder to train, collapsing at the default learning rate (\Cref{app:nano_setup}). With 1{,}000 support rows, about $7\times$ the pre-training task size, extra depth still does not help, although shallow models trained from scratch lead compressed ones by $0.016$--$0.027$ at $k = 2$ (\Cref{app:nano_context}).

\section{Conclusion}
\label{sec:conclusion}

\looseness=-1We introduced a closed-form, label-free affine translator to show that \glspl{tfm} are overparameterized. Across eight models from six families and 171 classification datasets, replacing half the transformer blocks with a single affine map costs at most $0.002$ median AUROC, and isolating a single block remains competitive on most datasets. Per-block analysis shows why this is possible: in single-axis models no individual block is critical, and in dual-axis models the non-affine computation sits in a single block, while attention remains diffuse across depth. Unlike early-exit methods that discard the network's tail and require labeled data to train a new classifier, our translator succeeds by preserving the pre-trained head and accurately projecting intermediate states into its expected input space. Furthermore, at half-depth, a fitted map can be reused as-is across similar datasets (e.g., MIMIC-III and eICU), and for TabICL and TabPFN-3 it keeps most of its benefit even on unrelated datasets. These reductions remove up to 96\% of the parameters with one block kept (1.5B for TabFM), speed up end-to-end GPU inference by up to $6\times$ at a median cost of $0.03$ AUROC, and let a single-block TabPFN-3 serve cached queries $2.4\times$ faster with $6\times$ less memory at negligible cost.

\paragraph{Limitations and future work.} Beyond demonstrating architectural redundancy, our analysis identifies specific conditions where extreme compression to a single block fails, such as maintaining continuous target rankings in regression, and for dual-axis models whose map cannot be absorbed (i.e., LimiX-16M), the per-feature map outweighs the removed blocks on very wide datasets. Additionally, our comparison between compression and shallow pre-training is currently limited to a 1.4M-parameter nanoTabPFN. Whether this performance parity holds at production scale, where pre-training shallow models from scratch is prohibitively expensive, remains an open question. These limitations motivate the design of adaptive architectures that allocate network depth only when strictly required. Given how replaceable most blocks are, a recurrent transformer that reuses one block's weights could supply that depth without the parameter count of current models. Finally, the window could be selected under a hardware budget instead of a fixed width, minimizing the calibration KL subject to a target memory or latency that also accounts for the cost of the map.

\bibliography{iclr2027_conference}
\bibliographystyle{iclr2027_conference}

\appendix
\raggedbottom 


\section{Implementation Details}
\label{app:setup}

\subsection{Datasets}
\label{app:datasets}

\Cref{tab:datasets} lists the classification and regression datasets. 

\begin{table}[h]
\centering
\caption{Dataset summary. Suite rows aggregate benchmarks deduplicated by OpenML ID, each dataset counted once under one suite; clinical rows list each dataset individually.}
\label{tab:datasets}
\small
\setlength{\tabcolsep}{4pt}
\begin{tabular}{lcccc}
\toprule
Name & \#Datasets & \#Instances & \#Features & \#Classes \\
\midrule
\multicolumn{5}{l}{\emph{Benchmark suites (classification)}} \\
\cmidrule{1-5}
OpenML-CC18 \citep{bischl2017openml} & 56 & 500--96K & 4--3{,}072 & 2--46 \\
TabZilla \citep{mcelfresh2023neural} & 44 & 148--1.0M & 4--4{,}296 & 2--100 \\
Grinsztajn \citep{grinsztajn2022tree} & 25 & 3.4K--940K & 7--419 & 2 \\
TabArena \citep{erickson2025tabarena} & 38 & 748--150K & 4--1{,}776 & 2--8 \\
\cmidrule{1-5}
\multicolumn{5}{l}{\emph{Clinical (classification)}} \\
\cmidrule{1-5}
Diabetes \citep{smith1988pima} & 1 & 768 & 8 & 2 \\
Heart \citep{statlog_heart_145} & 1 & 270 & 13 & 2 \\
Breast Cancer \citep{wolberg1992breast} & 1 & 699 & 9 & 2 \\
SUPPORT2 \citep{support1995controlled} & 1 & 9{,}105 & 46 & 2 \\
MIMIC-III \citep{johnson2016mimic} & 1 & 26{,}538 & 74 & 2 \\
eICU-CRD \citep{pollard2018eicu} & 1 & 30{,}000 & 74 & 2 \\
Cardiotocography \citep{cardiotocography_193} & 1 & 2{,}126 & 35 & 3 \\
Thyroid \citep{thyroid_disease_102} & 1 & 3{,}772 & 21 & 3 \\
\cmidrule{1-5}
\multicolumn{5}{l}{\emph{Benchmark suites (regression)}} \\
\cmidrule{1-5}
MIMIC-III length of stay \citep{johnson2016mimic} & 1 & 26{,}538 & 74 & --- \\
TabArena \citep{erickson2025tabarena} & 13 & 907--54K & 5--1{,}024 & --- \\
OpenML-CTR23 \citep{fischer2023ctr23} & 27 & 517--72K & 5--116 & --- \\
\cmidrule{1-5}
Total & 212 & & & \\
\bottomrule
\end{tabular}
\end{table}

We use 68/12/20\% train/calibration/test splits and three seeds (0, 1, 2) for every dataset, clinical and suite alike. Train, calibration and test splits are disjoint by index (test first, then calibration off the remainder), and patient-disjoint for the clinical cohorts (MIMIC-III, eICU-CRD) via \texttt{GroupShuffleSplit} on patient identifiers; encoders and scalers are fitted on the training rows only. A few standard benchmark datasets (e.g., Breast Cancer, Thyroid) contain a small number of intrinsically identical rows that can straddle splits by chance; these natural duplicates are exposed identically to the full and the compressed model, so they do not bias the relative compression cost.
Numerical features are standardised on the training set; categoricals are integer-encoded; missing numericals are imputed with zero after scaling. The calibration split is also the set on which the translator is fitted. Not every model runs on every dataset. TabPFNv2, TabFM, both Orions and LimiX-16M support at most 10 classes and skip the six suite datasets with more classes, and Orion-MSP and LimiX-16M run out of GPU memory on a few large high-dimensional datasets (Orion-MSP and Orion-Bix also miss single seeds on a few of them for the same reason). These models therefore run on 152--157 of the 163 suite datasets; all models run on the eight clinical datasets, and every comparison is between a model and its own full version on the same datasets.

\subsection{Models}
\label{app:models}

\subsubsection{GBDTs Hyperparameters}
\label{app:gbdt}

XGBoost and LightGBM are tuned per dataset and seed by a 30-configuration random search on the calibration split over a discrete grid --- \texttt{n\_estimators} $\in \{300, 500, 800, 1200\}$, \texttt{max\_depth} $\in \{4, 6, 8, 10\}$ (LightGBM: $\{-1, 6, 8, 12\}$, \texttt{num\_leaves} $\in \{31, 63, 127, 255\}$, \texttt{min\_child\_samples} $\in \{10, 20, 40\}$), \texttt{learning\_rate} $\in \{0.01, 0.03, 0.05, 0.1, 0.2\}$, \texttt{subsample} and \texttt{colsample\_bytree} $\in \{0.6, 0.8, 1.0\}$ --- with early stopping after 20 rounds, selected by calibration AUROC; they share test splits and seeds with the foundation-model rows. The default-hyperparameter variants are reported alongside in the suite rank tables.

\subsubsection{Tabular Foundation Models}
\label{app:tfm}

\Cref{tab:app_architecture} summarizes the architecture of each model.

\begin{table}[h]
    \centering
    \caption{Architecture of the eight evaluated models. The translator replaces blocks of the transformer stack only; encoders and head run unchanged. Params are the encoder and stack parameters (stack share of total), while context is the support-set cap in our experiments, not the model's limit.}
    \label{tab:app_architecture}
    \scriptsize
    \setlength{\tabcolsep}{3pt}
    \begin{tabularx}{\textwidth}{c c >{\raggedright\arraybackslash}X r l r c c}
        \toprule
                  &                 & \multicolumn{2}{c}{Encoders (never replaced)}                 & \multicolumn{2}{c}{Transformer stack (probed)} &                          &                  \\
        \cmidrule(lr){3-4} \cmidrule(lr){5-6}
        Model     & Tokenisation    & Encoder                                                       & Params (M)                                     & Blocks $\times$ $d$ & Params (share)   & Cached mode              & Context    \\
        \midrule
        TabPFNv2  & cell, dual-axis & linear ($x$ and $y$)                                          & 0.01                                           & 12 $\times$ 192     & 7.1 M (98\%)     & yes & 3{,}000 \\
        TabPFN-3  & row             & feature-distribution embedder + column aggregator             & 1.3                                            & 24 $\times$ 512     & 51.4 M (97\%)    & yes & 3{,}000 \\
        TabICL    & row             & set transformer per column + 3-layer row transformer          & 1.3                                            & 12 $\times$ 512     & 25.7 M (93\%)    & yes & 3{,}000 \\
        TabDPT    & row             & linear                                                        & 0.08                                           & 16 $\times$ 768     & 75.6 M (98\%)    & no (retrieval per query) & 1{,}500 \\
        TabFM     & row             & cell embedder + 2 column set transformers + 2 row interactors & 19.5                                           & 24 $\times$ 2048    & 1{,}611 M (98\%) & no & 3{,}000 \\
        Orion-MSP & row             & as TabICL + multi-scale row interactor + Perceiver memory     & 18.1                                           & 12 $\times$ 512     & 25.2 M (57\%)    & no & 3{,}000 \\
        Orion-Bix & row             & as TabICL                                                     & 1.3                                            & 12 $\times$ 512     & 25.2 M (93\%)    & no & 3{,}000 \\
        LimiX-16M & cell, dual-axis & mask-embedding linear                                         & 0.13                                           & 12 $\times$ 192     & 15.9 M (96\%)    & no & 3{,}000 \\
        \bottomrule
    \end{tabularx}
\end{table}

\paragraph{Ensembling.} Prediction-time ensembling averages independent forward passes, which would mix per-block residuals across passes, so we reduce it to a single member where possible. TabPFNv2, TabPFN-3, Orion and TabFM run with one estimator (library defaults: four, automatic, 32 and 32). TabICL (v2 checkpoint) is the exception: its default 8-member ensemble runs in a single forward pass, so we keep it. TabDPT runs at its default configuration and LimiX at its default with a single inference-configuration member. Full and compressed models use the same setting, so this choice does not affect any $\Delta$. To measure its cost, we refit each model with its library default on the eight clinical datasets, three seeds. Full AUROC rises by a mean of $0.002$ (TabPFNv2, TabPFN-3), $0.001$ (TabFM) and $0.006$ (Orion-MSP, Orion-Bix); medians are at most $0.003$. The largest gains are for the Orion models on the two ICU datasets ($+0.014$ to $+0.020$).

\paragraph{Effect of calibration rows on the full baseline.} The compressed model uses the calibration split to fit the translator and to select the window, thus the full model never sees it. To analyze the impact we refit every full model on train $\cup$ calibration (same support cap) and re-score the test split on the clinical datasets, three seeds. Only the five datasets below the cap can change (Diabetes, Heart, Breast Cancer, Cardiotocography, Thyroid): 117 runs (five datasets, eight models, three seeds; TabDPT on Thyroid already exceeds its 1{,}500-row cap). The median AUROC change is $0.000$ for every model, per-model means range from $-0.002$ to $+0.003$, and the 16 runs that move by more than $0.01$ split evenly (8 gains, 8 losses). Withholding the calibration rows does not impact the baseline.

\subsection{Translator implementation and ablations} \label{app:translator-family}

The following section provides additional details regarding \Cref{sec:method}.

\paragraph{Dual-axis models.} For the dual-axis models (i.e., TabPFNv2, LimiX-16M), block outputs have shape $(n_\text{items}, n_\text{features}, d)$ and the map for feature index $f$ is applied as $\mathbf{H}'[\cdot, f, \cdot] = \mathbf{H}[\cdot, f, \cdot]\, W_f + b_f$. With $n_\text{features} = 1$, the per-feature variant reduces exactly to the single-$(W,b)$ case, so this is a generalisation of the shared linear translator rather than a different method.  The number of feature tokens is set by each model's preprocessing (for LimiX, the frozen preprocessing below), so $n_\text{features}$ is the token count, not the column count.

\paragraph{LimiX.} LimiX refits its preprocessing on every \texttt{predict()} call on the concatenated rows, so the feature count and column identity drift between the calibration and evaluation passes, and a per-feature map would be applied to mismatched columns. We therefore fit the preprocessing once on the support rows and freeze it: categorical detection, the categorical encoder and the constant-column filter are fitted on the support set and only applied to the query rows, so the column layout is identical in every pass. With the frozen preprocessing the full model's predictions are identical on 7 of the 8 clinical datasets ($-0.006$ on Heart) and the per-feature map applies unchanged. On the 153 suite datasets both run, it loses $0.005$ on average at Half (median $0.002$, 83\% of datasets within one point) and $0.052$ at One-block (median $0.029$), against $0.003$ and $0.098$ for TabPFNv2. A shared $(W,b)$, which does not need the freezing, is near-lossless at width 3--4 but collapses with one block kept ($-0.151$ on the clinical datasets): the collapse reflects an incompatibility between the map and the preprocessing, not evidence that LimiX needs its depth.

\paragraph{Conditioning.} Each ridge solve has $d{+}1$ unknowns per output column. Its rows are sequence positions from one forward pass over the support and calibration sets, capped at 500 (all 216 on Heart, the smallest dataset).  The cap applies to positions only. TabPFNv2 and LimiX-16M fit one solve per feature token, each over the same $\leq 500$ positions. TabICL fits one $(W,b)$ across all 8 ensemble members (up to 4{,}000 correlated rows). The solve is therefore over-determined for TabPFNv2, LimiX-16M ($d{=}192$) and TabICL ($d{=}512$), slightly under-determined for TabPFN-3 and Orion ($d{=}512$), and under-determined for TabDPT ($d{=}768$) and TabFM ($d{=}2{,}048$). In the under-determined cases the ridge penalty makes the fit well-posed, and the training residual $\varepsilon$ becomes optimistic because the fit interpolates. All claims therefore rest on test AUROC, not on $\varepsilon$. The translator never sees test data or labels. Extraction and solve take 33--109~ms per window on one GPU.

\begin{figure}[h]
    \centering
    \includegraphics[width=\textwidth]{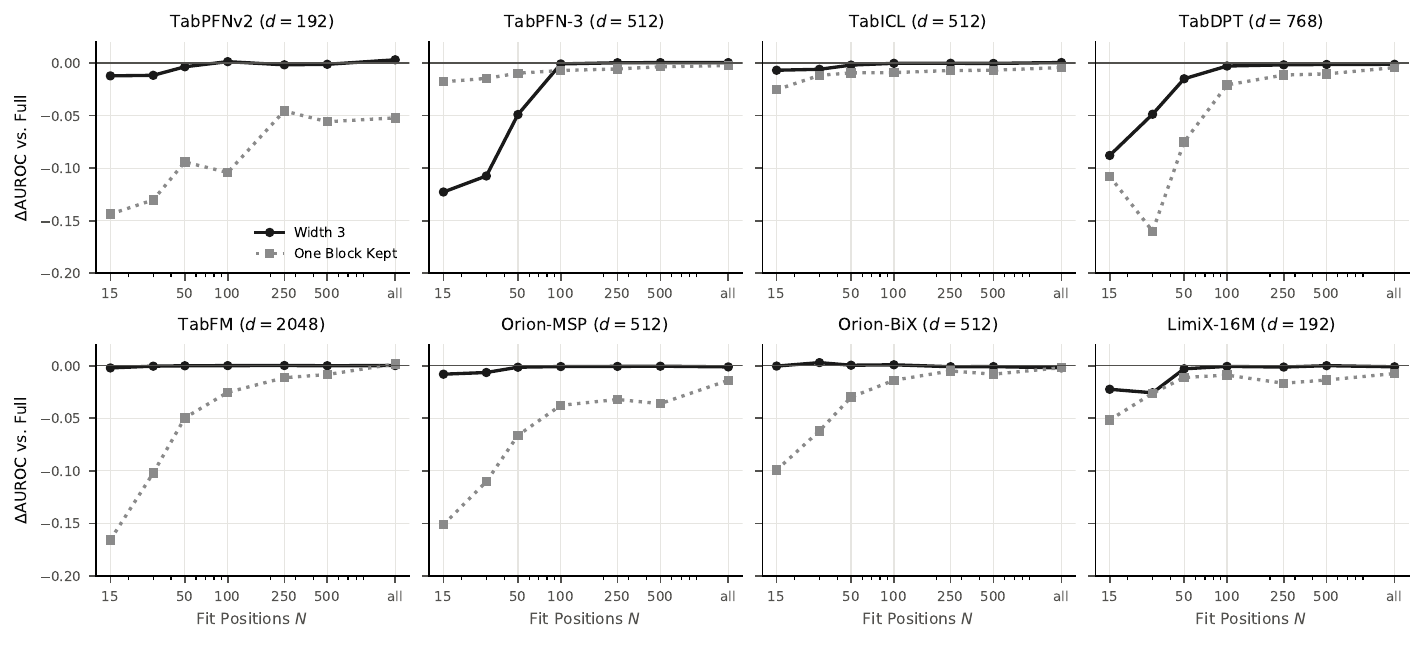}
    \caption{Calibration budget: $\Delta$AUROC vs.\ number of fitting positions $N$ (seed 0, five clinical datasets, three draws, KL-selected windows). Solid: width 3; dotted: one block kept. Mild compression needs at most 100 positions; with one block kept the knee moves to 250--500.}
    \label{fig:ncal}
\end{figure}

\paragraph{Calibration budget.} The translator is fitted on at most 500 positions of one forward pass. To test whether this is enough, we refit at the KL-selected windows on $N \in \{15, 30, 50, 100, 250, 500\}$ random subsets (three draws each, $\lambda$ chosen label-free as always) on five clinical datasets and compare with the capped fit. For mild compression the budget is small: at width 3 every model is within $0.005$ of the capped fit by $N{=}100$,  TabFM ($d{=}2{,}048$) and Orion-Bix from $N{=}15$, TabICL, TabPFNv2, Orion-MSP and LimiX-16M from $N{=}50$, TabPFN-3 and TabDPT from $N{=}100$,  because the KL-selected mild windows sit late in the stack, where representations change little. With one block kept the budget matters: TabDPT, TabFM and Orion-Bix need $N{=}250$ and TabPFNv2 the full 500, and at $N{=}15$ the one-block models fall $0.014$--$0.16$ below the capped fit. Fitting on all positions of the pass still helps Orion-MSP ($+0.022$) and TabFM ($+0.010$), and at most $0.006$ for the others. About 100 unlabelled positions therefore suffice for mild compression and 250--500 for extreme compression, far below the context the model already consumes (\Cref{fig:ncal}).

\paragraph{Per-feature vs.\ shared translator.} The per-feature translator has one $d \times d$ map per feature token, $37{,}056$ parameters per token for TabPFNv2, against $589{,}824$ per TabPFNv2 block; the map therefore exceeds the blocks it replaces once the dataset has more than $16w$ feature tokens for a window of width $w$ (64 tokens at width 4, 175 at one block kept). On MIMIC-III (39 tokens for 74 columns) the translator is already 61\% of the four blocks it replaces at width 4, although the net parameter change at one block kept remains $-71\%$; on the 14 suite datasets with more than 100 columns (61--302 tokens) it exceeds the four removed blocks on 12 of 14 and the eleven removed blocks on 6 of 14. A shared $d^2{+}d$ map would remove this dependence on the feature count, but it does not preserve accuracy (\Cref{tab:family}): at width 4 the per-feature map is within $0.02$ of the full model on every dataset (median $-0.002$), whereas the shared map loses up to $0.15$ (median $-0.014$; per-feature better on 10 of 14, worse on 1), and with one block kept the medians are $-0.082$ and $-0.301$ (per-feature better on 13 of 14). On wide datasets the dual-axis translator therefore measures redundancy but saves parameters only when the window starts at block $0$, where the map folds into the encoder; the single-axis models, where the map is a single $d^2{+}d$ regardless of the dataset, have no such limit (262k parameters for TabICL against 23.6M removed at one block kept).

\begin{table}[h]
    \centering
    \caption{Per-feature vs.\ shared translator for TabPFNv2 on 14 suite datasets.}
    \label{tab:family}
    \small
    \begin{tabular}{lrrrrrrr}
    \toprule
    & & & \multicolumn{1}{c}{translator} & \multicolumn{2}{c}{width 4} & \multicolumn{2}{c}{one block kept} \\
    \cmidrule(lr){5-6}\cmidrule(lr){7-8}
    Dataset & columns & tokens & params (M) & per-feat. & shared & per-feat. & shared \\
    \midrule
    cc18\_1485 & 500 & 252 & 9.3 & -0.018 & -0.152 & -0.321 & -0.427 \\
    grin\_45019 & 419 & 211 & 7.8 & +0.003 & -0.047 & -0.080 & -0.289 \\
    philippine & 308 & 156 & 5.8 & -0.009 & -0.091 & -0.154 & -0.261 \\
    cc18\_1501 & 256 & 184 & 6.8 & -0.001 & -0.021 & -0.085 & -0.416 \\
    cc18\_40979 & 240 & 302 & 11.2 & +0.000 & -0.007 & -0.056 & -0.408 \\
    cc18\_12 & 216 & 272 & 10.1 & -0.000 & -0.006 & -0.090 & -0.461 \\
    ta\_46939 & 212 & 106 & 3.9 & -0.013 & +0.013 & -0.047 & -0.269 \\
    cc18\_40670 & 180 & 92 & 3.4 & -0.001 & -0.003 & -0.012 & -0.472 \\
    volkert & 180 & 185 & 6.9 & -0.002 & -0.068 & -0.127 & -0.322 \\
    ta\_46908 & 170 & 86 & 3.2 & -0.001 & -0.001 & -0.103 & -0.012 \\
    jasmine & 144 & 74 & 2.7 & -0.016 & -0.025 & -0.024 & -0.085 \\
    tz\_40536 & 120 & 62 & 2.3 & -0.002 & -0.002 & -0.013 & -0.073 \\
    cc18\_1486 & 118 & 61 & 2.3 & -0.004 & -0.004 & -0.015 & -0.121 \\
    ta\_46980 & 111 & 101 & 3.7 & -0.003 & -0.056 & -0.131 & -0.313 \\
    \midrule
    median & & & & -0.002 & -0.014 & -0.082 & -0.301 \\
    \bottomrule
    \end{tabular}
\end{table}

\paragraph{Gradient refinement of the translator.} \label{app:refinement}
To test whether the closed-form translator can be improved with additional training, we refine it with 500 Adam steps of label-free final-representation matching through the frozen kept blocks. At width 4 this moves $\Delta$AUROC by at most $0.003$ and calibration KL by at most $0.005$ on any model (111 runs on the eight clinical datasets, three seeds); with one block kept the median change is $0.000$ and the largest $0.036$ (TabPFNv2 on Thyroid, where refinement moves further from Full). On regression at one block kept (MIMIC-III length of stay and the 13 TabArena regression sets, seed 0, $R^2$), refinement changes little for TabPFNv2, TabICL and TabFM (median gain $\leq 0.006$) and matters only for TabDPT, recovering a median $0.07$ of a median $0.45$ loss. The residual gap is therefore not an optimization artifact of the closed-form estimator: gradient descent on the same objective leaves it essentially unchanged, which attributes it to approximation error, i.e., to the part of the removed blocks' computation that no affine map of the kept representation reproduces. Supervised fine-tuning on the task loss, as in TACTICL, optimizes a different objective with labeled data and falls outside the training-free setting studied here.

\section{Additional Results} \label{app:additional_experiments}

\subsection{Attention analysis} \label{app:attention_entropy_def}

\paragraph{Measuring entropy.} We measure attention diversity via per-head entropy: $\mathcal{H}(\mathbf{a}) = -\sum_j a_j \log a_j$, averaged across heads, queries, and samples. An entropy near $\log(S)$ (where $S$ is the number of keys) signals uniform attention, while low entropy signals specialization. Because foundation models compute attention inside fused kernels that do not return weights, we recompute the softmax from the model's own projected queries and keys. We intercept the scaled-dot-product kernel during one forward pass and attribute each call to the block executing it. This interception includes the kernel's mask and scale, and applies TabFM's query-key normalization where relevant. We use the 500 unlabelled calibration rows as queries and all support rows as keys. We exclude the dual-axis models (TabPFNv2, LimiX) from this specific analysis because they interleave item and feature attention.

\paragraph{Top-$k$ overlap between blocks.}\label{app:attention_overlap}
During the same forward pass, we record the five keys with the largest attention weight for every head across the 500 calibration queries. We then compare blocks using the Jaccard index of these key sets. We evaluate this on MIMIC-III and Higgs (seed 0), yielding 3,000 candidate rows (1,500 for TabDPT).

Matching heads by index across blocks yields a Jaccard index of just $0.02$--$0.04$ for all evaluated models. Pooling the heads of a block (roughly 33 distinct keys per query) yields overall overlaps of $0.11$--$0.12$ (TabICL), $0.09$ (TabDPT), $0.11$--$0.12$ (TabFM), and $0.06$--$0.08$ (TabPFN-3). Consecutive blocks share slightly more ($0.10$--$0.19$), but this remains vastly different from the $0.005$ expected from random sets. Notably, block 0 in TabPFN-3 shares almost no keys ($0.01$) with later blocks. Ultimately, the attended neighborhoods change at every block.

\begin{figure}[h]
    \centering
    \includegraphics[width=\textwidth]{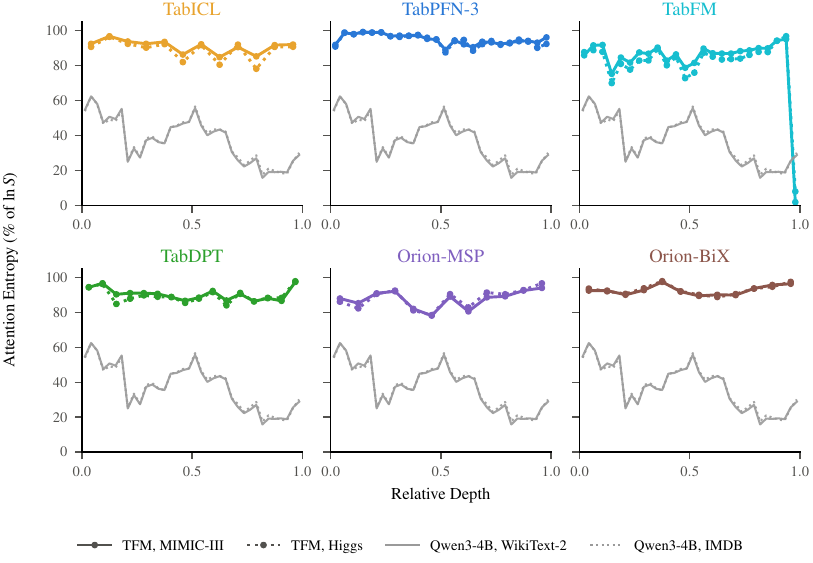}
    \caption{Per-block attention entropy as a percentage of the maximum possible entropy ($\ln S$) for all six tabular \glspl{tfm} on MIMIC-III (solid) and Higgs (dotted) vs.\ Qwen3-4B (dashed). All models maintain near-uniform attention (70--99\% outside TabFM's read-out block) compared to Qwen3-4B's sharply peaked attention (16--62\%). TabFM's read-out block drops to 2--8\%.}
    \label{fig:nlp_comparison_full}
\end{figure}

\paragraph{Comparison with a language model.}\label{app:nlp_comparison}
For the NLP comparison in \Cref{fig:nlp_comparison}, we evaluate six tabular \glspl{tfm} on MIMIC-III and Higgs (using 3,000 support rows; 1,500 for TabDPT). We compare these against Qwen3-4B~\citep{yang2025qwen3}, a 36-layer language model with grouped-query attention (32 query / 8 KV heads), evaluated on WikiText-2 and IMDB. This 2-vs-2 dataset design ensures the contrast is not an artifact of any single dataset. \Cref{fig:nlp_comparison_full} shows all six models; the main text shows the three most representative.

To compare models with different context lengths on one scale, we plot each block's entropy as a percentage of its maximum, $\ln S$ for $S$ attended keys. The difference is not in the absolute entropy, which scales with sequence length, but in its profile across depth. Tabular models stay near-uniform at every block, whereas Qwen3-4B starts high, dips around layers 7--9, recovers in the middle, and falls again in the last quarter.

\paragraph{Attention does not compensate for bypassed blocks.}\label{app:attention}
\begin{figure}[h]
    \centering
    \includegraphics[width=0.6\textwidth]{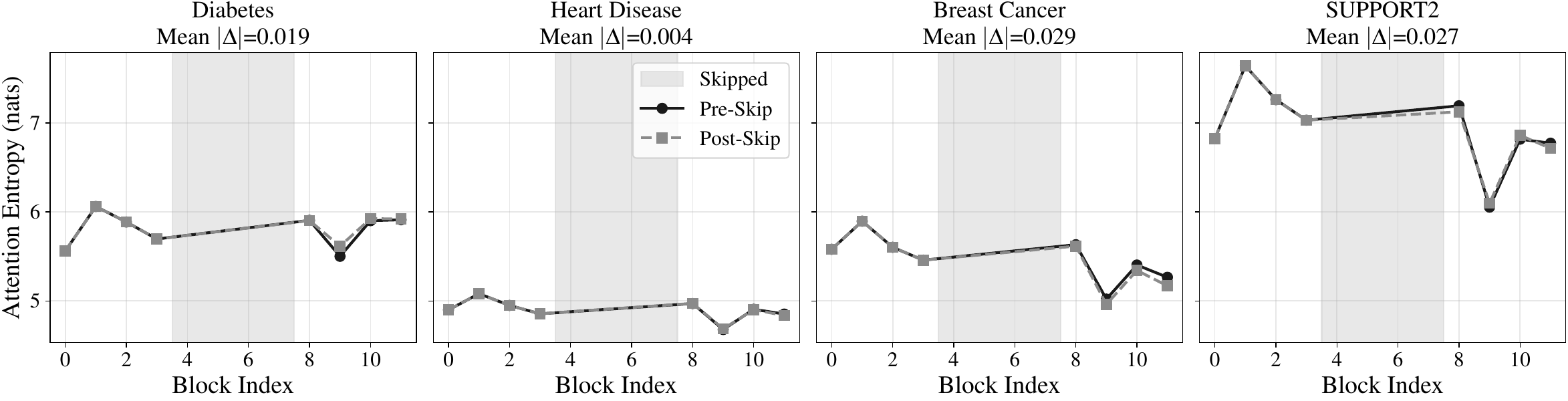}
    \caption{TabICL per-block attention entropy before vs.\ after replacing blocks 4--8 with the affine translator (shaded). Surviving blocks do not reorganise their attention to compensate.}
    \label{fig:attention}
\end{figure}

A direct test of the redundancy interpretation is whether surviving blocks reorganize their attention to compensate for the bypassed sub-stack. \Cref{fig:attention} shows the per-block mean attention entropy before and after replacing a canonical window (blocks 4--8) in TabICL. A detailed per-dataset breakdown is provided in \Cref{tab:pre_post_skip_entropy}.

\begin{table}[h]
    \centering
    \caption{TabICL attention stability: per-block entropy on surviving blocks before vs.\ after replacing a contiguous window. $\varepsilon$ = translator residual; $|\Delta\mathcal{H}|$ = absolute entropy change (nats). Shifts are $<0.03$ nats on average.}
    \label{tab:pre_post_skip_entropy}
    \small
    \setlength{\tabcolsep}{3pt}
    \begin{tabular}{lcccccc}
        \toprule
        Dataset          & Skip Window & $\varepsilon$ & Mean $|\Delta\mathcal{H}|$ & Max $|\Delta\mathcal{H}|$ & Pre $\overline{\mathcal{H}}$ & Post $\overline{\mathcal{H}}$ \\
        \midrule
        Diabetes         & 4--8        & .197          & .019                       & .114                      & 5.80                         & 5.82                          \\
        Heart            & 4--8        & .149          & .004                       & .017                      & 4.90                         & 4.90                          \\
        Br.\ Cancer      & 4--8        & .122          & .029                       & .097                      & 5.48                         & 5.45                          \\
        SUPPORT2         & 4--8        & .246          & .027                       & .068                      & 6.95                         & 6.94                          \\
        \midrule
        Diabetes (early) & 2--6        & .190          & .028                       & .110                      & 5.75                         & 5.76                          \\
        Heart (late)     & 6--10       & .097          & .001                       & .005                      & 4.89                         & 4.89                          \\
        \bottomrule
    \end{tabular}
\end{table}

Across datasets, the per-block entropy changes by less than $0.03$ nats on average, and at most $0.11$ nats on any single surviving block (roughly 1--2\% of the total entropy value). This indicates that the approximated sub-stack contributed negligibly to the downstream attention dynamics. Unsurprisingly, the shift is slightly larger for an early skip window (e.g., blocks 2--6 on Diabetes) because intervening closer to block 0 perturbs the input distribution to all subsequent blocks more heavily. Conversely, a late skip window (blocks 6--10 on Heart) causes almost zero shift (max $\Delta = 0.005$ nats). The surviving blocks therefore do not reorganize their attention, unlike the final layers of language transformers, which are thought to specialize to the training objective~\citep{dalvi2020analyzing}.

\subsection{Per-block sensitivity, all models}
\label{app:sensitivity_full}

\begin{figure}[h]
    \centering
    \includegraphics[width=\textwidth]{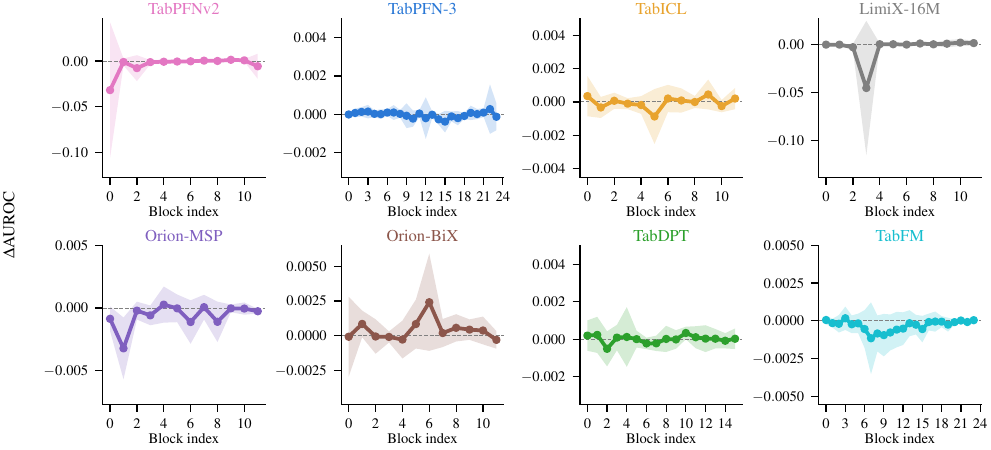}
    \caption{Per-block sensitivity of every reported model: $\Delta$AUROC when a single block is replaced by the ridge translator (mean over the eight clinical datasets and three seeds).}
    \label{fig:sensitivity_full}
\end{figure}

\Cref{fig:sensitivity_full} extends \Cref{fig:sensitivity} to every reported model.

\subsection{Evaluation}
\label{app:main_results}

\subsection{Window selection}\label{app:selection}

\paragraph{Window selection needs no labels.} Every classification window is the one whose compressed model is closest in KL to the full model on the calibration split. No label is needed (the ridge $\lambda$ is also chosen label-free), and because the head is kept, the KL costs nothing to compute. \Cref{tab:selection_criteria} compares this choice with the labelled one (best calibration AUROC), with the translator's residual $\varepsilon$, and with the test-set oracle. At Half, the KL matches or beats the labelled choice for every model except Orion-Bix ($-0.001$ vs.\ $+0.000$). $\varepsilon$ never beats the KL: it is worse for six models and ties for the two Orions, because it ignores how the head amplifies errors and is not comparable across depths. At One-block, where the only choice is keeping the first or the last block, the single-axis models stay within $0.002$ of the labelled choice; for the two dual-axis models, the KL sometimes keeps the costlier block (\Cref{tab:kept_block}). 

\begin{table}[h]
\centering
\caption{Window selection on the full sweep. Mean over datasets of the test $\Delta$AUROC (averaged over seeds) (compressed $-$ full) of the window chosen by each criterion, per (model, dataset, seed): the smallest calibration KL $\mathrm{KL}(p_\text{full}\,\|\,p_\text{c})$ (ours, no labels; every result in the paper), the highest calibration AUROC (labelled), the smallest translator fit residual $\varepsilon$, and the highest test AUROC (oracle, an upper bound). All suite and clinical datasets the model runs on ($n$), three seeds, ridge translators.}
\label{tab:selection_criteria}
\scriptsize
\setlength{\tabcolsep}{4pt}
\resizebox{\textwidth}{!}{%
\begin{tabular}{lrrrrrrrrr}
\toprule
 & \multicolumn{4}{c}{Half: mean $\Delta$AUROC} & \multicolumn{4}{c}{One-block: mean $\Delta$AUROC} & \\
\cmidrule(lr){2-5} \cmidrule(lr){6-9}
Model & KL (ours) & Labelled & $\varepsilon$ & Oracle & KL (ours) & Labelled & $\varepsilon$ & Oracle & $n$ \\
\midrule
TabPFN-3 & $-$.001 & $-$.001 & $-$.003 & $+$.001 & $-$.006 & $-$.006 & $-$.007 & $-$.005 & 171 \\
TabICL & $-$.003 & $-$.003 & $-$.004 & $-$.001 & $-$.015 & $-$.015 & $-$.016 & $-$.014 & 171 \\
TabDPT & $-$.003 & $-$.004 & $-$.005 & $-$.000 & $-$.024 & $-$.025 & $-$.025 & $-$.023 & 171 \\
TabFM & $-$.004 & $-$.005 & $-$.007 & $-$.002 & $-$.025 & $-$.026 & $-$.025 & $-$.025 & 165 \\
Orion-MSP & $-$.002 & $-$.003 & $-$.002 & $+$.001 & $-$.018 & $-$.018 & $-$.018 & $-$.016 & 160 \\
Orion-Bix & $-$.001 & $+$.000 & $-$.001 & $+$.004 & $-$.008 & $-$.008 & $-$.008 & $-$.006 & 165 \\
TabPFNv2 & $-$.003 & $-$.005 & $-$.010 & $-$.000 & $-$.098 & $-$.073 & $-$.097 & $-$.070 & 165 \\
LimiX-16M & $-$.005 & $-$.005 & $-$.011 & $+$.000 & $-$.051 & $-$.037 & $-$.111 & $-$.034 & 161 \\
\bottomrule
\end{tabular}}
\end{table}

\begin{table}[h]
\centering
\caption{Half depth needs no window selection for the single-axis models. Mean over the suite datasets of the $\Delta$AUROC (averaged over seeds) at Half (width $L/2$) with the window chosen by the calibration KL (no labels), with two fixed windows that need no selection at all (replace the last half of the stack; keep block 0 and replace the next $L/2$ blocks), and with the test-set oracle as the bound; in parentheses the share of datasets losing more than one point. Three seeds, ridge translators. TabPFNv2 (per-feature map) needs the selection.}
\label{tab:fixed_window}
\scriptsize
\setlength{\tabcolsep}{4pt}
\resizebox{\textwidth}{!}{%
\begin{tabular}{lrrrrr}
\toprule
Model & KL-selected & Fixed: last half & Fixed: keep block 0, replace next $L/2$ & Test oracle & $n$ \\
\midrule
TabPFN-3 & $-$.001 (4\%) & $-$.003 (13\%) & $-$.007 (26\%) & $+$.001 & 163 \\
TabICL & $-$.003 (10\%) & $-$.004 (17\%) & $-$.008 (27\%) & $-$.001 & 163 \\
TabDPT & $-$.003 (9\%) & $-$.006 (23\%) & $-$.009 (31\%) & $-$.000 & 163 \\
TabFM & $-$.005 (13\%) & $-$.006 (13\%) & $-$.007 (23\%) & $-$.002 & 157 \\
Orion-MSP & $-$.002 (7\%) & $-$.002 (7\%) & $-$.012 (33\%) & $+$.000 & 152 \\
Orion-Bix & $-$.001 (3\%) & $-$.001 (4\%) & $-$.005 (20\%) & $+$.004 & 157 \\
TabPFNv2 & $-$.003 (9\%) & $-$.029 (64\%) & $-$.124 (79\%) & $-$.000 & 157 \\
\bottomrule
\end{tabular}}
\end{table}

\paragraph{KL-selected depth.}\label{app:kl_frontier} The calibration KL can also choose how many blocks to replace. \Cref{tab:kl_frontier} applies the tolerance rule of \Cref{eq:tolerance} ($\epsilon = 0.01$) on all 171 datasets, over the five swept widths, and on the eight clinical datasets, growing one window a block at a time. On all 171 datasets it removes 52--70\% of the blocks of the single-axis models (42\% for TabPFNv2, 22\% for LimiX-16M) at a mean cost of $0.001$--$0.004$ AUROC, with 85--97\% of datasets within one point. Block by block on the clinical datasets it removes more for all models but TabPFNv2 (70--81\% for the single-axis models, 50\% for LimiX-16M, but 32\% for TabPFNv2) at a similar cost, although greedy growth does not always find the lowest-KL window of each width.

\begin{table}[h]
\centering
\caption{Label-free depth from a drift tolerance. For each dataset and seed, the deepest compression whose calibration KL to the full model stays at most 0.01: on the 171 suite and clinical datasets among the swept widths $\{3, 4, L/2, L{-}2, L{-}1\}$ (KL-selected window per width), and on the eight clinical datasets along the greedy path that grows one window block by block. Removed = mean share of blocks replaced; $\Delta$AUROC = mean test cost (compressed $-$ full); $\geq -0.01$ = share of datasets within one point; greedy = min-KL = share of path steps where the greedy window is the lowest-KL window of that width. No label is used at any step. Three seeds.}
\label{tab:kl_frontier}
\scriptsize
\setlength{\tabcolsep}{3pt}
\resizebox{\textwidth}{!}{%
\begin{tabular}{lrrrrrrr}
\toprule
 & \multicolumn{3}{c}{All datasets (5 widths)} & \multicolumn{3}{c}{Clinical (8 datasets, greedy)} & \\
\cmidrule(lr){2-4} \cmidrule(lr){5-7}
Model & removed & $\Delta$AUROC & $\geq -0.01$ & removed & $\Delta$AUROC & $\geq -0.01$ & greedy = min-KL \\
\midrule
TabPFN-3 & 63\% & -0.001 & 95\% & 70\% & -0.001 & 100\% & 22\% \\
TabICL & 61\% & -0.003 & 94\% & 81\% & -0.002 & 100\% & 64\% \\
TabDPT & 52\% & -0.003 & 91\% & 74\% & -0.000 & 100\% & 46\% \\
TabFM & 53\% & -0.002 & 95\% & 79\% & -0.003 & 75\% & 81\% \\
Orion-MSP & 60\% & -0.002 & 93\% & 73\% & -0.001 & 100\% & 92\% \\
Orion-Bix & 70\% & -0.004 & 85\% & 73\% & +0.003 & 100\% & 74\% \\
TabPFNv2 & 42\% & -0.002 & 95\% & 32\% & +0.001 & 100\% & 65\% \\
LimiX-16M & 22\% & -0.001 & 97\% & 50\% & -0.000 & 100\% & 56\% \\
\bottomrule
\end{tabular}
}
\end{table}

\subsection{Classification results}\label{app:classification}

\begin{figure}[h]
    \centering
    \includegraphics[width=\textwidth]{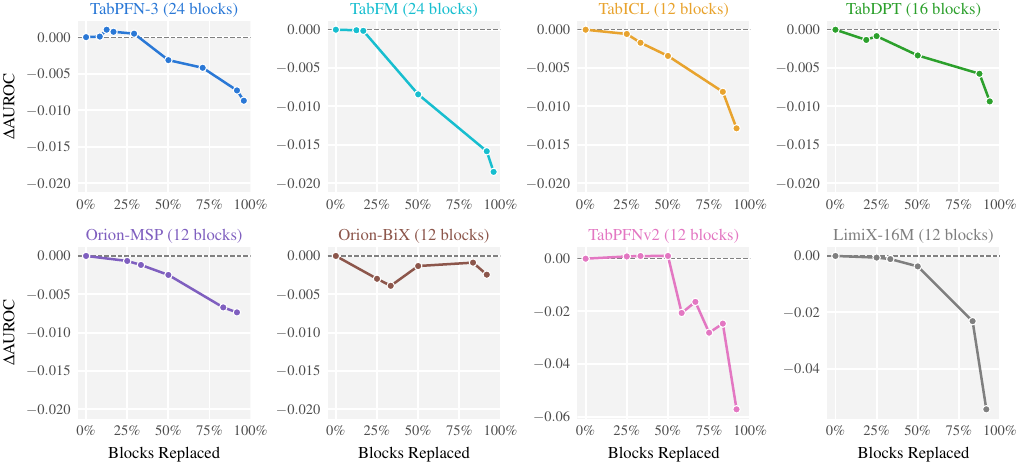}
    \caption{$\Delta$AUROC vs.\ share of blocks replaced on MIMIC-III for every model, at every swept width (three seeds, KL-selected windows, ridge translator).}
    \label{fig:compression_per_model}
\end{figure}

\Cref{fig:compression_per_model} extends the left panel of \Cref{fig:teaser} to every model.

\begin{table*}[t]
\centering
\caption{Classification results (AUROC). Clinical and benchmark suites, mean $\pm$ std over datasets and 3 seeds. Half: $L/2$ blocks replaced. One-block: $L{-}1$ blocks replaced. Ridge translators; windows selected by calibration KL (no labels). Id: the One-block window with the translator replaced by the identity (naive dropping). Per-dataset clinical results in \Cref{tab:foundation}.}
\label{tab:compact}
\scriptsize
\setlength{\tabcolsep}{3pt}
\begin{tabularx}{\textwidth}{l l *{5}{>{\centering\arraybackslash}X}}
\toprule
Model & & Clinical (8) & TabArena & TabZilla & CC-18 & Grinsztajn \\
\midrule
XGBoost & & .911$_{\pm.084}$ & .860$_{\pm.100}$ & .907$_{\pm.086}$ & .911$_{\pm.123}$ & .856$_{\pm.096}$ \\
LightGBM & & .909$_{\pm.086}$ & .859$_{\pm.100}$ & .907$_{\pm.086}$ & .911$_{\pm.124}$ & .857$_{\pm.097}$ \\
\midrule
\multirow{4}{*}{TabPFNv2}
 & Full & .904$_{\pm.099}$ & .852$_{\pm.112}$ & .901$_{\pm.091}$ & .913$_{\pm.122}$ & .834$_{\pm.099}$ \\
 & Half (6/12) & .904$_{\pm.097}$ & .850$_{\pm.113}$ & .896$_{\pm.093}$ & .911$_{\pm.122}$ & .829$_{\pm.099}$ \\
 & One-block (11/12) & .862$_{\pm.103}$ & .792$_{\pm.113}$ & .793$_{\pm.137}$ & .777$_{\pm.158}$ & .758$_{\pm.091}$ \\
 & Id (11/12) & .481$_{\pm.083}$ & .503$_{\pm.097}$ & .489$_{\pm.080}$ & .521$_{\pm.070}$ & .498$_{\pm.065}$ \\
\midrule
\multirow{4}{*}{TabPFN-3}
 & Full & .908$_{\pm.093}$ & .861$_{\pm.105}$ & .907$_{\pm.089}$ & .913$_{\pm.122}$ & .841$_{\pm.098}$ \\
 & Half (12/24) & .907$_{\pm.093}$ & .860$_{\pm.105}$ & .905$_{\pm.089}$ & .913$_{\pm.122}$ & .839$_{\pm.098}$ \\
 & One-block (23/24) & .906$_{\pm.094}$ & .856$_{\pm.105}$ & .898$_{\pm.094}$ & .910$_{\pm.123}$ & .834$_{\pm.098}$ \\
 & Id (23/24) & .880$_{\pm.112}$ & .808$_{\pm.105}$ & .823$_{\pm.115}$ & .866$_{\pm.130}$ & .775$_{\pm.099}$ \\
\midrule
\multirow{4}{*}{TabICL}
 & Full & .912$_{\pm.089}$ & .863$_{\pm.105}$ & .914$_{\pm.087}$ & .919$_{\pm.118}$ & .844$_{\pm.096}$ \\
 & Half (6/12) & .912$_{\pm.090}$ & .860$_{\pm.104}$ & .910$_{\pm.087}$ & .918$_{\pm.118}$ & .837$_{\pm.098}$ \\
 & One-block (11/12) & .909$_{\pm.092}$ & .852$_{\pm.102}$ & .891$_{\pm.089}$ & .910$_{\pm.120}$ & .822$_{\pm.097}$ \\
 & Id (11/12) & .839$_{\pm.098}$ & .768$_{\pm.109}$ & .763$_{\pm.132}$ & .773$_{\pm.130}$ & .745$_{\pm.091}$ \\
\midrule
\multirow{4}{*}{TabDPT}
 & Full & .902$_{\pm.097}$ & .846$_{\pm.113}$ & .887$_{\pm.101}$ & .905$_{\pm.125}$ & .823$_{\pm.100}$ \\
 & Half (8/16) & .903$_{\pm.097}$ & .843$_{\pm.113}$ & .881$_{\pm.105}$ & .904$_{\pm.126}$ & .819$_{\pm.100}$ \\
 & One-block (15/16) & .898$_{\pm.098}$ & .821$_{\pm.106}$ & .855$_{\pm.101}$ & .887$_{\pm.132}$ & .799$_{\pm.101}$ \\
 & Id (15/16) & .485$_{\pm.071}$ & .496$_{\pm.096}$ & .481$_{\pm.115}$ & .497$_{\pm.082}$ & .550$_{\pm.083}$ \\
\midrule
\multirow{4}{*}{TabFM}
 & Full & .914$_{\pm.088}$ & .867$_{\pm.102}$ & .926$_{\pm.081}$ & .921$_{\pm.116}$ & .875$_{\pm.089}$ \\
 & Half (12/24) & .912$_{\pm.090}$ & .865$_{\pm.103}$ & .919$_{\pm.086}$ & .920$_{\pm.118}$ & .865$_{\pm.086}$ \\
 & One-block (23/24) & .908$_{\pm.092}$ & .852$_{\pm.103}$ & .891$_{\pm.089}$ & .910$_{\pm.119}$ & .817$_{\pm.109}$ \\
 & Id (23/24) & .799$_{\pm.097}$ & .688$_{\pm.105}$ & .699$_{\pm.122}$ & .710$_{\pm.117}$ & .685$_{\pm.098}$ \\
\midrule
\multirow{4}{*}{Orion-MSP}
 & Full & .883$_{\pm.100}$ & .837$_{\pm.112}$ & .872$_{\pm.108}$ & .899$_{\pm.133}$ & .813$_{\pm.106}$ \\
 & Half (6/12) & .880$_{\pm.099}$ & .834$_{\pm.111}$ & .871$_{\pm.106}$ & .897$_{\pm.135}$ & .810$_{\pm.106}$ \\
 & One-block (11/12) & .870$_{\pm.094}$ & .823$_{\pm.108}$ & .848$_{\pm.108}$ & .886$_{\pm.134}$ & .795$_{\pm.105}$ \\
 & Id (11/12) & .755$_{\pm.146}$ & .732$_{\pm.133}$ & .722$_{\pm.129}$ & .770$_{\pm.117}$ & .741$_{\pm.116}$ \\
\midrule
\multirow{4}{*}{Orion-Bix}
 & Full & .886$_{\pm.111}$ & .823$_{\pm.129}$ & .858$_{\pm.123}$ & .889$_{\pm.143}$ & .796$_{\pm.123}$ \\
 & Half (6/12) & .888$_{\pm.110}$ & .820$_{\pm.134}$ & .859$_{\pm.123}$ & .889$_{\pm.142}$ & .795$_{\pm.124}$ \\
 & One-block (11/12) & .891$_{\pm.106}$ & .810$_{\pm.140}$ & .851$_{\pm.119}$ & .883$_{\pm.140}$ & .789$_{\pm.123}$ \\
 & Id (11/12) & .775$_{\pm.151}$ & .704$_{\pm.132}$ & .701$_{\pm.146}$ & .737$_{\pm.148}$ & .692$_{\pm.139}$ \\
\midrule
\multirow{4}{*}{LimiX-16M}
 & Full & .904$_{\pm.095}$ & .855$_{\pm.108}$ & .902$_{\pm.099}$ & .914$_{\pm.125}$ & .842$_{\pm.095}$ \\
 & Half (6/12) & .904$_{\pm.095}$ & .852$_{\pm.110}$ & .895$_{\pm.102}$ & .912$_{\pm.125}$ & .831$_{\pm.100}$ \\
 & One-block (11/12) & .888$_{\pm.109}$ & .823$_{\pm.105}$ & .826$_{\pm.114}$ & .857$_{\pm.132}$ & .806$_{\pm.102}$ \\
 & Id (11/12) & .565$_{\pm.130}$ & .515$_{\pm.097}$ & .498$_{\pm.088}$ & .510$_{\pm.092}$ & .499$_{\pm.084}$ \\
\bottomrule
\end{tabularx}
\end{table*}

\subsection{Per-dataset results and the identity control}
\label{app:foundation_tables}

\Cref{tab:foundation} gives the per-dataset clinical results behind the clinical column of \Cref{tab:compact}, with the \textbf{Id} rows: the One-block window with the translator replaced by the identity (naive layer dropping), which isolates the translator's contribution.

\begin{table*}[t]
\centering
\caption{Clinical datasets. AUROC for tuned GBDT baselines and eight tabular foundation models. \textbf{Full} = pre-trained model unchanged; Half = $L/2$ blocks replaced by the translator; One-block = $L{-}1$ of $L$ blocks replaced; Id = the same maximal window with the translator replaced by the identity (naive layer dropping), isolating the translator's contribution. Translators: ridge with label-free $\lambda$ selection; all skip windows selected label-free by the smallest calibration-split KL to the full model; subscripts $\pm$ std over seeds, where each seed is a different train/calibration/test split (patient-disjoint on MIMIC-III and eICU) and support subsample, so the Full rows vary with the seed as well. GBDTs are tuned (30-config random search on the calibration split) and share test splits and seeds with the foundation-model rows. All entries are raw AUROC.}
\label{tab:foundation}
\scriptsize
\setlength{\tabcolsep}{2pt}
\begin{tabularx}{\textwidth}{l l *{8}{>{\centering\arraybackslash}X}}
\toprule
Model & & Diabetes & Heart & Br. Cancer & SUPPORT2 & MIMIC & eICU & CTG$^\dagger$ & Thyroid$^\dagger$ \\
\midrule
XGBoost & & .828$_{\pm.022}$ & .885$_{\pm.041}$ & .988$_{\pm.008}$ & .981$_{\pm.001}$ & .819$_{\pm.005}$ & .792$_{\pm.002}$ & .997$_{\pm.001}$ & 1.00$_{\pm.000}$ \\
LightGBM & & .819$_{\pm.012}$ & .880$_{\pm.036}$ & .990$_{\pm.007}$ & .981$_{\pm.001}$ & .817$_{\pm.009}$ & .789$_{\pm.004}$ & .997$_{\pm.001}$ & 1.00$_{\pm.000}$ \\
\midrule
\multirow{4}{*}{TabPFNv2}
 & Full     & .844$_{\pm.024}$ & .900$_{\pm.030}$ & .995$_{\pm.004}$ & .980$_{\pm.000}$ & .773$_{\pm.011}$ & .741$_{\pm.004}$ & .998$_{\pm.000}$ & 1.00$_{\pm.000}$ \\
 & Half (6/12) & .841$_{\pm.027}$ & .903$_{\pm.036}$ & .994$_{\pm.003}$ & .976$_{\pm.001}$ & .774$_{\pm.014}$ & .750$_{\pm.003}$ & .996$_{\pm.001}$ & .998$_{\pm.001}$ \\
 & One-block (11/12) & .826$_{\pm.021}$ & .842$_{\pm.047}$ & .995$_{\pm.003}$ & .931$_{\pm.005}$ & .716$_{\pm.018}$ & .708$_{\pm.007}$ & .979$_{\pm.008}$ & .901$_{\pm.014}$ \\
 & Id (11/12) & .624$_{\pm.047}$ & .314$_{\pm.082}$ & .547$_{\pm.026}$ & .503$_{\pm.010}$ & .452$_{\pm.007}$ & .455$_{\pm.004}$ & .466$_{\pm.023}$ & .488$_{\pm.064}$ \\
\midrule
\multirow{4}{*}{TabPFN-3}
 & Full     & .842$_{\pm.028}$ & .901$_{\pm.029}$ & .995$_{\pm.005}$ & .979$_{\pm.001}$ & .792$_{\pm.010}$ & .759$_{\pm.005}$ & .998$_{\pm.002}$ & .999$_{\pm.000}$ \\
 & Half (12/24) & .841$_{\pm.028}$ & .899$_{\pm.032}$ & .995$_{\pm.005}$ & .977$_{\pm.001}$ & .789$_{\pm.009}$ & .760$_{\pm.005}$ & .998$_{\pm.001}$ & .999$_{\pm.000}$ \\
 & One-block (23/24) & .840$_{\pm.023}$ & .905$_{\pm.032}$ & .995$_{\pm.005}$ & .975$_{\pm.003}$ & .783$_{\pm.009}$ & .758$_{\pm.004}$ & .998$_{\pm.002}$ & .999$_{\pm.001}$ \\
 & Id (23/24) & .797$_{\pm.022}$ & .911$_{\pm.037}$ & .991$_{\pm.007}$ & .942$_{\pm.005}$ & .715$_{\pm.011}$ & .715$_{\pm.006}$ & .974$_{\pm.009}$ & .993$_{\pm.002}$ \\
\midrule
\multirow{4}{*}{TabICL}
 & Full     & .845$_{\pm.027}$ & .904$_{\pm.035}$ & .994$_{\pm.004}$ & .983$_{\pm.001}$ & .800$_{\pm.008}$ & .771$_{\pm.007}$ & .999$_{\pm.001}$ & 1.00$_{\pm.000}$ \\
 & Half (6/12) & .846$_{\pm.027}$ & .908$_{\pm.037}$ & .994$_{\pm.005}$ & .982$_{\pm.001}$ & .797$_{\pm.008}$ & .769$_{\pm.007}$ & .999$_{\pm.000}$ & 1.00$_{\pm.000}$ \\
 & One-block (11/12) & .844$_{\pm.023}$ & .911$_{\pm.038}$ & .994$_{\pm.004}$ & .975$_{\pm.001}$ & .788$_{\pm.010}$ & .764$_{\pm.005}$ & .999$_{\pm.000}$ & .998$_{\pm.000}$ \\
 & Id (11/12) & .799$_{\pm.030}$ & .897$_{\pm.042}$ & .988$_{\pm.007}$ & .923$_{\pm.002}$ & .708$_{\pm.024}$ & .692$_{\pm.033}$ & .893$_{\pm.013}$ & .809$_{\pm.102}$ \\
\midrule
\multirow{4}{*}{TabDPT}
 & Full     & .841$_{\pm.027}$ & .898$_{\pm.032}$ & .994$_{\pm.005}$ & .969$_{\pm.003}$ & .772$_{\pm.003}$ & .750$_{\pm.006}$ & .996$_{\pm.005}$ & .997$_{\pm.002}$ \\
 & Half (8/16) & .844$_{\pm.026}$ & .900$_{\pm.035}$ & .995$_{\pm.003}$ & .968$_{\pm.003}$ & .769$_{\pm.005}$ & .752$_{\pm.006}$ & .998$_{\pm.002}$ & .996$_{\pm.002}$ \\
 & One-block (15/16) & .842$_{\pm.030}$ & .908$_{\pm.036}$ & .995$_{\pm.004}$ & .953$_{\pm.005}$ & .763$_{\pm.006}$ & .738$_{\pm.012}$ & .995$_{\pm.002}$ & .989$_{\pm.002}$ \\
 & Id (15/16) & .432$_{\pm.031}$ & .611$_{\pm.014}$ & .442$_{\pm.027}$ & .456$_{\pm.013}$ & .505$_{\pm.012}$ & .471$_{\pm.003}$ & .578$_{\pm.011}$ & .386$_{\pm.017}$ \\
\midrule
\multirow{4}{*}{TabFM}
 & Full     & .844$_{\pm.020}$ & .909$_{\pm.028}$ & .992$_{\pm.006}$ & .984$_{\pm.001}$ & .811$_{\pm.009}$ & .771$_{\pm.006}$ & .999$_{\pm.001}$ & 1.00$_{\pm.000}$ \\
 & Half (12/24) & .840$_{\pm.024}$ & .912$_{\pm.029}$ & .992$_{\pm.004}$ & .983$_{\pm.001}$ & .802$_{\pm.008}$ & .765$_{\pm.006}$ & .998$_{\pm.003}$ & 1.00$_{\pm.000}$ \\
 & One-block (23/24) & .840$_{\pm.026}$ & .913$_{\pm.025}$ & .992$_{\pm.004}$ & .975$_{\pm.001}$ & .792$_{\pm.010}$ & .757$_{\pm.008}$ & .998$_{\pm.001}$ & .995$_{\pm.001}$ \\
 & Id (23/24) & .707$_{\pm.033}$ & .864$_{\pm.045}$ & .971$_{\pm.026}$ & .885$_{\pm.006}$ & .689$_{\pm.010}$ & .690$_{\pm.010}$ & .819$_{\pm.038}$ & .766$_{\pm.010}$ \\
\midrule
\multirow{4}{*}{Orion-MSP}
 & Full     & .831$_{\pm.036}$ & .870$_{\pm.053}$ & .995$_{\pm.005}$ & .888$_{\pm.007}$ & .756$_{\pm.005}$ & .733$_{\pm.005}$ & .999$_{\pm.000}$ & .995$_{\pm.002}$ \\
 & Half (6/12) & .830$_{\pm.035}$ & .871$_{\pm.055}$ & .994$_{\pm.006}$ & .875$_{\pm.005}$ & .753$_{\pm.003}$ & .732$_{\pm.005}$ & .999$_{\pm.001}$ & .987$_{\pm.001}$ \\
 & One-block (11/12) & .832$_{\pm.033}$ & .875$_{\pm.049}$ & .995$_{\pm.004}$ & .859$_{\pm.004}$ & .748$_{\pm.004}$ & .730$_{\pm.007}$ & .995$_{\pm.000}$ & .928$_{\pm.007}$ \\
 & Id (11/12) & .809$_{\pm.036}$ & .856$_{\pm.038}$ & .987$_{\pm.005}$ & .498$_{\pm.029}$ & .673$_{\pm.026}$ & .659$_{\pm.012}$ & .881$_{\pm.029}$ & .676$_{\pm.019}$ \\
\midrule
\multirow{4}{*}{Orion-Bix}
 & Full     & .778$_{\pm.044}$ & .870$_{\pm.014}$ & .993$_{\pm.006}$ & .973$_{\pm.002}$ & .752$_{\pm.007}$ & .727$_{\pm.002}$ & .998$_{\pm.000}$ & .997$_{\pm.002}$ \\
 & Half (6/12) & .790$_{\pm.041}$ & .876$_{\pm.013}$ & .994$_{\pm.005}$ & .972$_{\pm.002}$ & .751$_{\pm.008}$ & .727$_{\pm.004}$ & .998$_{\pm.000}$ & .996$_{\pm.002}$ \\
 & One-block (11/12) & .816$_{\pm.031}$ & .901$_{\pm.032}$ & .994$_{\pm.005}$ & .958$_{\pm.004}$ & .750$_{\pm.008}$ & .722$_{\pm.006}$ & .998$_{\pm.000}$ & .988$_{\pm.001}$ \\
 & Id (11/12) & .772$_{\pm.023}$ & .873$_{\pm.042}$ & .988$_{\pm.006}$ & .618$_{\pm.196}$ & .547$_{\pm.058}$ & .619$_{\pm.006}$ & .885$_{\pm.010}$ & .896$_{\pm.025}$ \\
\midrule
\multirow{4}{*}{LimiX-16M}
 & Full     & .837$_{\pm.033}$ & .872$_{\pm.017}$ & .995$_{\pm.005}$ & .982$_{\pm.001}$ & .789$_{\pm.007}$ & .760$_{\pm.006}$ & .999$_{\pm.001}$ & .999$_{\pm.000}$ \\
 & Half (6/12) & .836$_{\pm.021}$ & .881$_{\pm.035}$ & .994$_{\pm.005}$ & .979$_{\pm.001}$ & .785$_{\pm.006}$ & .760$_{\pm.004}$ & .999$_{\pm.000}$ & .999$_{\pm.000}$ \\
 & One-block (11/12) & .842$_{\pm.020}$ & .885$_{\pm.010}$ & .988$_{\pm.004}$ & .977$_{\pm.001}$ & .735$_{\pm.055}$ & .707$_{\pm.031}$ & .980$_{\pm.005}$ & .988$_{\pm.004}$ \\
 & Id (11/12) & .541$_{\pm.023}$ & .540$_{\pm.061}$ & .875$_{\pm.022}$ & .528$_{\pm.005}$ & .588$_{\pm.073}$ & .510$_{\pm.115}$ & .550$_{\pm.012}$ & .385$_{\pm.021}$ \\
\bottomrule
\multicolumn{10}{l}{\scriptsize $^\dagger$Macro-averaged OVR AUROC (CTG, Thyroid 3-class).} \\
\end{tabularx}
\end{table*}

\subsection{Suite-scale rank tables} \label{app:suite_ranks}

\Cref{tab:suite_ranks} ranks every model over the suite datasets, with its full and Half versions ranked together (left) and the Half versions against tuned GBDTs (right). Replacing half the stack leaves the ordering of the models unchanged: the Half models rank in the same order as the full ones. The Half TabFM (average rank 3.53) is outranked only by the full TabFM (2.26) and beats every other full model. Against the trees, the Half versions of TabFM, TabICL and TabPFN-3 all rank above tuned LightGBM and XGBoost. Removing half the depth therefore changes neither which model is best nor how the models compare with trees.

\begin{table}[t]
\centering
\caption{Average rank over the shared suite datasets (lower is better, $\pm$ std; Friedman $p < 10^{-100}$). Left: Full and Half versions jointly ranked. Right: Half models vs.\ tuned GBDTs. Half-depth TabFM outranks every other full model.}
\label{tab:suite_ranks}
\scriptsize
\setlength{\tabcolsep}{2pt}
\begin{minipage}[t]{0.54\textwidth}\centering
\resizebox{\linewidth}{!}{%
\begin{tabular}{lrrrr}
\toprule
Method & Avg.\ rank & $\pm$ & Best & Mean AUROC \\
\midrule
TabFM & 2.26 & 2.48 & 101 & 0.9017 \\
TabFM (Half) & 3.53 & 2.54 & 6 & 0.8971 \\
TabICL & 3.88 & 2.35 & 10 & 0.8912 \\
TabICL (Half) & 5.29 & 2.26 & 4 & 0.8883 \\
TabPFN-3 & 5.54 & 2.58 & 5 & 0.8868 \\
TabPFN-3 (Half) & 6.26 & 2.64 & 6 & 0.8858 \\
LimiX-16M & 7.74 & 3.38 & 1 & 0.8845 \\
TabPFNv2 & 7.95 & 2.91 & 3 & 0.8814 \\
LimiX-16M (Half) & 9.67 & 2.95 & 2 & 0.8794 \\
TabDPT & 9.80 & 3.10 & 3 & 0.8720 \\
TabPFNv2 (Half) & 9.86 & 3.00 & 1 & 0.8781 \\
TabDPT (Half) & 11.50 & 2.95 & 0 & 0.8686 \\
Orion-MSP & 12.46 & 2.78 & 1 & 0.8622 \\
Orion-Bix & 13.23 & 2.89 & 1 & 0.8488 \\
Orion-MSP (Half) & 13.31 & 2.61 & 0 & 0.8603 \\
Orion-Bix (Half) & 13.72 & 2.86 & 0 & 0.8484 \\
\bottomrule
\end{tabular}
}
\end{minipage}\hfill
\begin{minipage}[t]{0.44\textwidth}\centering
\resizebox{\linewidth}{!}{%
\begin{tabular}{lrrrr}
\toprule
Method & Avg.\ rank & $\pm$ & Best & Mean AUROC \\
\midrule
TabFM (Half) & 2.63 & 1.84 & 66 & 0.8971 \\
TabICL (Half) & 3.88 & 2.07 & 17 & 0.8883 \\
TabPFN-3 (Half) & 4.61 & 2.37 & 13 & 0.8858 \\
LightGBM (tuned) & 5.29 & 3.35 & 26 & 0.8882 \\
XGBoost (tuned) & 5.44 & 3.17 & 11 & 0.8882 \\
LimiX-16M (Half) & 6.91 & 2.44 & 2 & 0.8794 \\
LightGBM & 6.96 & 3.16 & 1 & 0.8820 \\
TabPFNv2 (Half) & 7.02 & 2.59 & 2 & 0.8781 \\
XGBoost & 7.73 & 3.17 & 2 & 0.8802 \\
TabDPT (Half) & 8.19 & 2.94 & 0 & 0.8686 \\
Orion-MSP (Half) & 9.59 & 2.49 & 1 & 0.8603 \\
Orion-Bix (Half) & 9.76 & 2.78 & 2 & 0.8484 \\
\bottomrule
\end{tabular}
}
\end{minipage}
\end{table}

\Cref{tab:suite_wtl} counts, per model, the datasets on which the compressed version wins, ties or loses against its own full version (a tie is $|\Delta\text{AUROC}| \leq 0.001$).

\begin{table}[h]
    \centering
    \caption{Per-model win/tie/loss of the compressed model (Half window, ridge) against its own full version over the suite datasets, with the median $\Delta$AUROC.}
    \label{tab:suite_wtl}
    \small
\begin{tabular}{lrrrr}
\toprule
Model & Win & Tie & Loss & Median $\Delta$AUROC \\
\midrule
TabPFNv2 & 14 & 48 & 95 & -0.0017 \\
TabPFN-3 & 19 & 77 & 67 & -0.0004 \\
TabICL & 7 & 70 & 86 & -0.0011 \\
TabDPT & 13 & 55 & 95 & -0.0018 \\
TabFM & 10 & 47 & 100 & -0.0021 \\
Orion-MSP & 22 & 44 & 86 & -0.0014 \\
Orion-Bix & 27 & 72 & 58 & -0.0004 \\
LimiX-16M & 23 & 33 & 97 & -0.0019 \\
\bottomrule
\end{tabular}
\end{table}

\subsection{Check against TabArena} \label{app:reference_check}
Since our protocol differs from the models' own evaluations (a single ensemble member where possible, a support set capped at 3{,}000 rows (1{,}500 for TabDPT), a 68/12/20 split per seed, and no bagging), in \Cref{tab:reference_check} we compare our Full AUROC (mean over three seeds) with the default-configuration entries of the public TabArena leaderboard~\citep{erickson2025tabarena} on the 30 binary TabArena datasets, where the metric is the same and every dataset matches by name. TabArena evaluates each configuration on 3-fold outer cross-validation (repeated ten times below 2{,}500 rows, three times above) with the full training portion and cross-validation ensembling. Our numbers track the leaderboard per dataset (Pearson $r$ $0.93$--$0.97$) and sit $0.4$--$1.2$ points below it on the datasets with at most 3{,}000 rows, where only the split protocol and ensembling differ, and $0.7$--$2.9$ points below on the larger datasets, where our support cap also binds. Neither gap affects any claim in the paper, which compares each model with its own Full baseline under an identical protocol.

\begin{table}[h]
    \centering
    \caption{Our Full baselines vs.\ the TabArena leaderboard (default configurations, ROC AUC, 30 binary TabArena datasets): per-dataset Pearson correlation and the mean difference, split at 3{,}000 rows.}
    \label{tab:reference_check}
    \small
    \begin{tabular}{lrrrrrr}
    \toprule
    & & & \multicolumn{2}{c}{$\leq$ 3,000 rows} & \multicolumn{2}{c}{$>$ 3,000 rows} \\
    \cmidrule(lr){4-5}\cmidrule(lr){6-7}
    Model & datasets & Pearson $r$ & $n$ & ours $-$ TabArena & $n$ & ours $-$ TabArena \\
    \midrule
    TabPFN-3 & 30 & 0.951 & 9 & -0.004 & 21 & -0.022 \\
    TabICL & 30 & 0.962 & 9 & -0.005 & 21 & -0.018 \\
    TabDPT & 30 & 0.935 & 9 & -0.004 & 21 & -0.029 \\
    TabFM & 30 & 0.968 & 9 & -0.007 & 21 & -0.018 \\
    LimiX-16M & 30 & 0.933 & 9 & -0.012 & 21 & -0.007 \\
    \bottomrule
    \end{tabular}
\end{table}

\subsection{Decision metrics beyond AUROC} \label{app:decision_metrics}

AUROC measures how well a model ranks examples, so we also check the decisions the compressed models make. \Cref{tab:acc_suite} reports accuracy and balanced accuracy at both operating points. At Half, the compressed models make essentially the same decisions as the full ones: both metrics stay within about $0.01$ of the full model on binary tasks and within $0.02$ on multi-class tasks. With one block kept, an extreme setting, the decision metrics drop more than AUROC, mostly on multi-class tasks, where a small shift between the probabilities of similar classes can change the predicted class without changing the ranking. TabPFN-3 is the most robust (at most $0.03$ balanced accuracy lost), and the largest drops come from the two dual-axis models, which are also the most affected in AUROC.

\begin{table}[h]
\centering
\caption{AUROC, accuracy and balanced accuracy at both operating points. Mean over datasets of Full / Half / One-block (KL-selected windows, ridge translator, three seeds; decisions by the most probable class, no recalibration), for the binary and the multi-class datasets (suite and clinical); $n$ = datasets. At Half every metric stays close to Full; with one block kept the threshold-dependent metrics fall more than AUROC, most on multi-class tasks.}
\label{tab:acc_suite}
\small
\setlength{\tabcolsep}{3pt}
\begin{tabular}{lrccccccccc}
\toprule
& & \multicolumn{3}{c}{AUROC} & \multicolumn{3}{c}{Accuracy} & \multicolumn{3}{c}{Balanced accuracy} \\
\cmidrule(lr){3-5}\cmidrule(lr){6-8}\cmidrule(lr){9-11}
Model & $n$ & Full & Half & One & Full & Half & One & Full & Half & One \\
\midrule
\multicolumn{11}{l}{\emph{Binary tasks}} \\
TabPFNv2 & 114 & .858 & .856 & .797 & .840 & .837 & .771 & .753 & .749 & .626 \\
TabPFN-3 & 114 & .865 & .864 & .860 & .846 & .844 & .839 & .761 & .756 & .746 \\
TabICL & 114 & .868 & .865 & .853 & .849 & .846 & .832 & .765 & .760 & .733 \\
TabDPT & 114 & .849 & .847 & .828 & .833 & .830 & .813 & .741 & .734 & .691 \\
TabFM & 114 & .880 & .876 & .854 & .859 & .854 & .829 & .780 & .767 & .712 \\
Orion-MSP & 111 & .838 & .836 & .822 & .824 & .822 & .807 & .730 & .725 & .682 \\
Orion-Bix & 114 & .827 & .826 & .819 & .818 & .818 & .811 & .720 & .717 & .694 \\
LimiX-16M & 113 & .864 & .858 & .831 & .845 & .839 & .800 & .762 & .750 & .662 \\
\midrule
\multicolumn{11}{l}{\emph{Multi-class tasks}} \\
TabPFNv2 & 51 & .928 & .925 & .748 & .826 & .817 & .479 & .765 & .747 & .334 \\
TabPFN-3 & 57 & .940 & .938 & .932 & .844 & .840 & .825 & .789 & .781 & .759 \\
TabICL & 57 & .947 & .944 & .932 & .862 & .855 & .817 & .811 & .803 & .747 \\
TabDPT & 57 & .929 & .924 & .897 & .823 & .813 & .725 & .764 & .750 & .620 \\
TabFM & 51 & .945 & .940 & .922 & .867 & .857 & .802 & .812 & .794 & .707 \\
Orion-MSP & 49 & .925 & .923 & .904 & .803 & .795 & .746 & .743 & .728 & .642 \\
Orion-Bix & 51 & .896 & .895 & .889 & .780 & .778 & .753 & .705 & .700 & .651 \\
LimiX-16M & 48 & .929 & .926 & .837 & .843 & .835 & .580 & .779 & .766 & .449 \\
\bottomrule
\end{tabular}
\end{table}

\subsection{Subgroup analysis on the clinical cohorts} \label{app:subgroups}
Does compression move the predictions of some patients more than others? We compare the per-row test predictions of the full and the compressed model on MIMIC-III and eICU (patient-disjoint splits, three seeds) at Half and One-block, at the KL-selected windows, along two axes a clinical deployment cares about: the completeness of a patient's record (the fraction of the 74 raw features observed before imputation; quartiles of the test rows, Q1 the least complete) and the outcome (\Cref{tab:subgroups}). Decisions are compared at a common threshold, the full model's Youden-$J$ threshold on the calibration split, so that a change in sensitivity or specificity reflects the compressed model's probabilities and not a re-tuned cut-off. At Half the change is small and uniform: for the six single-axis models and the two dual-axis models the $\Delta$AUROC of every completeness quartile is within $0.007$ of the overall value, sensitivity on the patients who died and specificity on those who survived move by at most $0.07$ for the single-axis models ($0.05$ for TabPFNv2), and the mean absolute change of the predicted probability is $0.01$--$0.06$ for the patients who died against $0.005$--$0.03$ for those who survived. At One-block the cost lands where the tail of \Cref{sec:main-results} predicts. Across completeness quartiles the single-axis models stay within $0.02$ of their overall cost, while TabPFNv2 loses most on the \emph{most} complete records (Q4 $-0.048$ against Q1 $-0.015$ on eICU, $-0.057$ against $-0.035$ on MIMIC-III). Across outcomes, the probabilities of the patients who died move about $2.5\times$ more than those of the survivors ($0.03$--$0.10$ against $0.02$--$0.04$ for TabICL, TabFM, TabPFN-3 and TabDPT), and at the common threshold this is a loss of sensitivity at a gain of specificity: $-0.17$ and $-0.12$ (TabFM), $-0.03$ and $-0.08$ (TabICL), $-0.02$ and $-0.05$ (TabPFN-3) on eICU and MIMIC-III, against $+0.01$ to $+0.09$ specificity; the Orions and TabDPT move by at most $0.05$. TabPFNv2's One-block predictions shift wholesale on MIMIC-III (sensitivity $+0.24$, specificity $-0.64$), a calibration change rather than a ranking one. LimiX-16M's One-block model loses $0.054$ AUROC on both cohorts and its probabilities fall below the full model's threshold, costing $0.28$--$0.37$ sensitivity for $0.11$--$0.19$ specificity. The redundancy is therefore not paid for by a subgroup at Half, and at One-block the patients it costs are the ones with the rarer outcome, which is the clinical argument for Half as the deployable point. The outcome is the minority class (11--14\% of rows).

\begin{table}[h]
\centering
\caption{Subgroup analysis on MIMIC-III and eICU (patient-disjoint, 3 seeds, windows selected by calibration KL). Per model at Half and One-block: overall $\Delta$AUROC; $\Delta$AUROC by feature-completeness quartile (Q1 = least complete); $\Delta$sensitivity/$\Delta$specificity at the full model's Youden-$J$ threshold; mean $|\Delta p|$ by outcome group.}
\label{tab:subgroups}
\scriptsize
\setlength{\tabcolsep}{3pt}
\resizebox{\textwidth}{!}{%
\begin{tabular}{llrrrrrrr|rrrrrrr}
\toprule
 & & \multicolumn{7}{c|}{Half} & \multicolumn{7}{c}{One-block} \\
\cmidrule(lr){3-9}\cmidrule(lr){10-16}
Model & Cohort & $\Delta$AUROC & Q1 & Q4 & $\Delta$sens & $\Delta$spec & $|\Delta p|$ died & surv. & $\Delta$AUROC & Q1 & Q4 & $\Delta$sens & $\Delta$spec & $|\Delta p|$ died & surv. \\
\midrule
\multirow{2}{*}{TabPFNv2} & MIMIC-III & $+0.001$ & $+0.000$ & $+0.004$ & $+0.001$ & $-0.011$ & 0.053 & 0.029 & $-0.057$ & $-0.035$ & $-0.057$ & $+0.243$ & $-0.637$ & 0.132 & 0.093 \\
 & eICU & $+0.008$ & $+0.013$ & $+0.007$ & $-0.048$ & $+0.036$ & 0.058 & 0.028 & $-0.033$ & $-0.015$ & $-0.048$ & $-0.015$ & $-0.166$ & 0.107 & 0.073 \\
\midrule
\multirow{2}{*}{TabPFN-3} & MIMIC-III & $-0.003$ & $+0.002$ & $-0.007$ & $-0.022$ & $+0.019$ & 0.027 & 0.014 & $-0.009$ & $-0.002$ & $-0.014$ & $-0.048$ & $+0.042$ & 0.043 & 0.023 \\
 & eICU & $+0.002$ & $-0.001$ & $+0.003$ & $-0.020$ & $+0.018$ & 0.022 & 0.010 & $-0.001$ & $-0.006$ & $-0.001$ & $-0.024$ & $+0.015$ & 0.031 & 0.016 \\
\midrule
\multirow{2}{*}{TabICL} & MIMIC-III & $-0.004$ & $-0.003$ & $-0.004$ & $+0.006$ & $-0.009$ & 0.019 & 0.010 & $-0.012$ & $-0.005$ & $-0.012$ & $-0.077$ & $+0.038$ & 0.063 & 0.025 \\
 & eICU & $-0.003$ & $-0.001$ & $-0.008$ & $+0.013$ & $-0.014$ & 0.016 & 0.008 & $-0.011$ & $-0.007$ & $-0.016$ & $-0.031$ & $+0.006$ & 0.048 & 0.023 \\
\midrule
\multirow{2}{*}{TabDPT} & MIMIC-III & $-0.003$ & $-0.002$ & $-0.001$ & $-0.014$ & $-0.002$ & 0.025 & 0.013 & $-0.009$ & $-0.008$ & $-0.010$ & $-0.031$ & $+0.016$ & 0.064 & 0.027 \\
 & eICU & $+0.002$ & $-0.005$ & $+0.001$ & $+0.007$ & $-0.006$ & 0.029 & 0.015 & $-0.013$ & $-0.027$ & $-0.016$ & $-0.021$ & $-0.012$ & 0.075 & 0.030 \\
\midrule
\multirow{2}{*}{TabFM} & MIMIC-III & $-0.008$ & $-0.006$ & $-0.010$ & $-0.049$ & $+0.023$ & 0.045 & 0.018 & $-0.019$ & $-0.012$ & $-0.024$ & $-0.122$ & $+0.063$ & 0.096 & 0.035 \\
 & eICU & $-0.006$ & $-0.009$ & $-0.007$ & $-0.071$ & $+0.037$ & 0.036 & 0.015 & $-0.014$ & $-0.020$ & $-0.017$ & $-0.173$ & $+0.087$ & 0.075 & 0.027 \\
\midrule
\multirow{2}{*}{Orion-MSP} & MIMIC-III & $-0.002$ & $-0.000$ & $-0.002$ & $-0.015$ & $+0.009$ & 0.028 & 0.014 & $-0.007$ & $-0.008$ & $-0.004$ & $-0.045$ & $+0.027$ & 0.058 & 0.027 \\
 & eICU & $-0.001$ & $-0.005$ & $-0.002$ & $+0.017$ & $-0.011$ & 0.025 & 0.013 & $-0.003$ & $-0.021$ & $-0.007$ & $-0.025$ & $+0.020$ & 0.045 & 0.023 \\
\midrule
\multirow{2}{*}{Orion-Bix} & MIMIC-III & $-0.001$ & $+0.001$ & $-0.002$ & $+0.009$ & $-0.008$ & 0.015 & 0.008 & $-0.002$ & $+0.008$ & $+0.002$ & $+0.010$ & $-0.007$ & 0.020 & 0.011 \\
 & eICU & $-0.000$ & $+0.003$ & $-0.002$ & $+0.005$ & $-0.006$ & 0.007 & 0.005 & $-0.005$ & $-0.006$ & $-0.004$ & $+0.008$ & $-0.020$ & 0.009 & 0.007 \\
\midrule
\multirow{2}{*}{LimiX-16M} & MIMIC-III & $-0.004$ & $-0.004$ & $-0.005$ & $-0.031$ & $+0.026$ & 0.047 & 0.025 & $-0.054$ & $-0.030$ & $-0.052$ & $-0.365$ & $+0.192$ & 0.175 & 0.069 \\
 & eICU & $-0.001$ & $-0.007$ & $-0.000$ & $-0.004$ & $-0.005$ & 0.035 & 0.018 & $-0.054$ & $-0.108$ & $-0.062$ & $-0.282$ & $+0.107$ & 0.110 & 0.052 \\
\bottomrule
\end{tabular}}
\end{table}

\subsection{Ablations: Early exit vs.\ translator}
\label{app:early_exit}

Since an affine map suffices to bridge the removed blocks, one could ask whether the class is already linearly decodable at the window start, so that the later blocks and the head add nothing a linear readout could not. We test this with an early-exit probe: the later blocks and the pre-trained head are discarded, and a logistic regression is fitted on the intermediate representations $\mathbf{H}_s$ with the support labels. If decodability at $s$ were the whole story, this supervised probe should match the label-free translator. However, across 2{,}664 configurations over seven models (all but LimiX-16M) on the eight clinical datasets, the translator matches or beats the probe in 81\% of cases. The probe's median cost stays near $-0.02$ at every depth ($-0.023$ early, $-0.019$ late), while the translator's shrinks from $-0.003$ to $+0.000$. The two exceptions are consistent with \Cref{sec:main-results}: TabPFNv2's probe wins at early exits ($-0.048$ vs.\ $-0.082$), and Orion-Bix has the lowest win rate (63\%) because its probe is already near-lossless ($-0.010$ to $-0.005$). The later blocks therefore do not refine an already-linear decision boundary; they move representations into the coordinates the pre-trained head reads, and the translator reproduces that move without labels.

\begin{table}[h]
    \centering
    \caption{Early-exit probe vs.\ affine translator: share of configurations where the translator matches or beats the probe, and median $\Delta$AUROC vs.\ the full model across three depth bands (8 clinical datasets, 3 seeds, all widths). $L$ = total blocks; $\ell$ = exit layer / translator start.}
    \label{tab:early_exit_summary}
    \scriptsize
    \setlength{\tabcolsep}{4pt}
\begin{tabular}{lrr|rr|rr|rr}
\toprule
& & & \multicolumn{2}{c|}{early ($\ell \leq L/3$)} & \multicolumn{2}{c|}{mid} & \multicolumn{2}{c}{late ($\ell > 2L/3$)} \\
Model & runs & transl.\ $\geq$ probe & probe & transl. & probe & transl. & probe & transl. \\
\midrule
TabPFNv2 & 288 & 67\% & -0.048 & -0.082 & -0.035 & -0.003 & -0.016 & +0.000 \\
TabPFN-3 & 576 & 77\% & -0.017 & -0.002 & -0.015 & -0.002 & -0.014 & -0.000 \\
TabICL & 288 & 95\% & -0.022 & -0.001 & -0.024 & -0.000 & -0.018 & +0.000 \\
TabDPT & 384 & 86\% & -0.032 & -0.003 & -0.032 & -0.002 & -0.030 & -0.000 \\
TabFM & 552 & 91\% & -0.026 & -0.004 & -0.018 & -0.002 & -0.031 & -0.000 \\
Orion-MSP & 288 & 80\% & -0.050 & -0.006 & -0.041 & -0.002 & -0.031 & +0.000 \\
Orion-Bix & 288 & 63\% & -0.010 & -0.003 & -0.005 & -0.001 & -0.006 & +0.000 \\
\midrule
\textbf{All} & 2664 & 81\% & -0.023 & -0.003 & -0.022 & -0.002 & -0.019 & +0.000 \\
\bottomrule
\end{tabular}
\end{table}

\subsection{One-block Ablations} \label{app:kept_block}

\paragraph{Which block is kept.} At One-block, the window keeps either the first or the last block, and the calibration KL chooses between them (\Cref{tab:kept_block}). The single-axis models mostly keep the first block, which is also the cheaper option for them. TabPFN-3 costs the same either way, since its encoders do most of the work. The dual-axis models mostly keep the last block: keeping the first costs TabPFNv2 $0.281$ and LimiX-16M $0.230$ AUROC, against $0.071$ and $0.034$ for the last. In the 8--25\% of cases where the KL keeps their first block, it picks the costlier option.

\paragraph{Looping the kept block.} To test whether the kept block contributes computation or only bridges to the head, we apply it $r \in \{1, 2, 3, 6, 12\}$ times, with the translator refitted on the looped output or removed (\Cref{tab:looped_block}). With the translator, repetition brings no gain: the cost stays within $0.003$ for TabICL and $0.006$ for TabDPT, and grows from $0.006$ to $0.016$ for TabFM. Without it, no number of repetitions recovers the model (TabDPT stays near $-0.4$). The affine map, not additional computation from the kept block, is what makes One-block work.

\begin{table}[h]
\centering
\caption{Which block is kept at One-block. Share of dataset--seed pairs whose KL-selected window keeps the first block rather than the last (suite / clinical datasets), and mean $\Delta$AUROC when the first or the last block is kept (averaged over seeds; $n$: datasets with both windows).}
\label{tab:kept_block}
\small
\setlength{\tabcolsep}{4pt}

\begin{tabular}{lrrrr}
\toprule
Model & Keeps first block (suite / clinical) & Cost, first kept & Cost, last kept & $n$ \\
\midrule
TabPFNv2 & 14\% / 8\% & $-$.281 & $-$.071 & 165 \\
TabPFN-3 & 43\% / 92\% & $-$.007 & $-$.006 & 171 \\
TabICL & 86\% / 79\% & $-$.016 & $-$.019 & 171 \\
TabDPT & 97\% / 100\% & $-$.025 & $-$.049 & 171 \\
TabFM & 93\% / 96\% & $-$.025 & $-$.036 & 165 \\
Orion-MSP & 92\% / 79\% & $-$.017 & $-$.023 & 160 \\
Orion-Bix & 83\% / 83\% & $-$.008 & $-$.012 & 165 \\
LimiX-16M & 16\% / 25\% & $-$.230 & $-$.034 & 161 \\
\bottomrule
\end{tabular}
\end{table}

\begin{table}[h]
\centering
\caption{Looping the kept block $r$ times, with the translator refitted on its output (+ map) or removed (no map). $\Delta$AUROC vs.\ the full model, eight clinical datasets, three seeds.}
\label{tab:looped_block}
\small
\setlength{\tabcolsep}{4pt}

\begin{tabular}{llrrrrr}
\toprule
Model & Variant & $r=1$ & $r=2$ & $r=3$ & $r=6$ & $r=12$ \\
\midrule
TabICL & kept block $\times r$ + map & $-$.002 & $-$.002 & $-$.002 & $-$.003 & $-$.003 \\
 & kept block $\times r$, no map & $-$.067 & $-$.034 & $-$.034 & $-$.044 & $-$.071 \\
\midrule
TabDPT & kept block $\times r$ + map & $-$.005 & $-$.005 & $-$.004 & $-$.005 & $-$.006 \\
 & kept block $\times r$, no map & $-$.417 & $-$.409 & $-$.430 & $-$.480 & $-$.404 \\
\midrule
TabFM & kept block $\times r$ + map & $-$.006 & $-$.014 & $-$.015 & $-$.016 & $-$.016 \\
 & kept block $\times r$, no map & $-$.113 & $-$.047 & $-$.051 & $-$.085 & $-$.179 \\
\bottomrule
\end{tabular}
\end{table}

\subsection{Context size}
\label{app:context}

\begin{figure}[h]
    \centering
    \includegraphics[width=\textwidth]{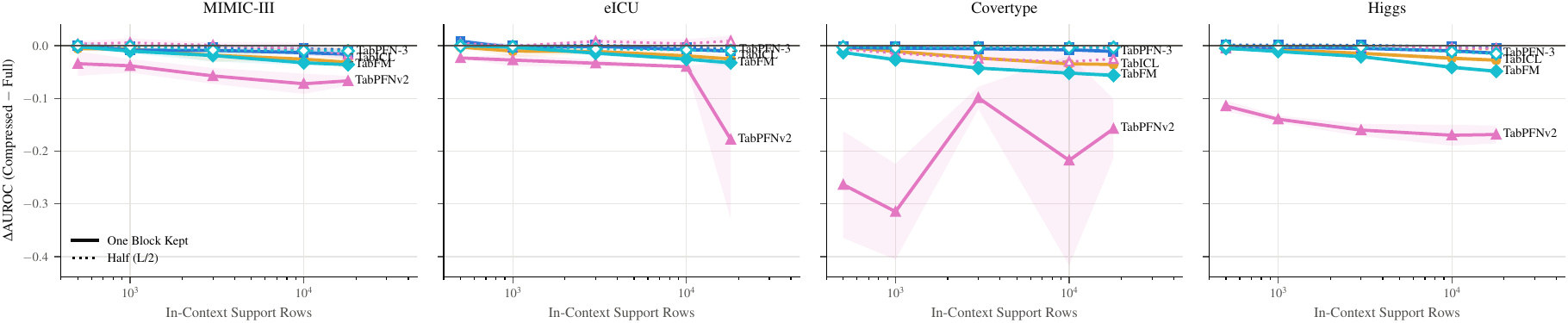}
    \caption{Context size vs.\ compression cost: $\Delta$AUROC as the support set grows from 500 to 18{,}000 rows (3 seeds, shaded $\pm$ std; windows re-selected at every size). Solid: one block kept; dotted: Half ($L/2$). At Half the mean cost stays within $0.009$ at every size (worst dataset: TabPFNv2 on Covertype, $-0.030$); with one block kept it grows with context (TabFM: $-0.005$ at 500 rows to $-0.043$ at 18{,}000).}
    \label{fig:context}
\end{figure}

\Cref{fig:context} shows the context-size experiment behind \Cref{sec:main-results}. Across the suite, the same trend appears with dataset size: larger datasets lose more at One-block (Spearman $\rho=-0.22$ between the number of rows and the One-block cost, over the 163 suite datasets and eight models, $p < 10^{-14}$; $-0.09$ (LimiX-16M) to $-0.34$ per model).

\subsection{Zero-shot map transfer across datasets}
\label{sec:transfer}

\begin{figure}[h]
    \centering
    \includegraphics[width=\columnwidth]{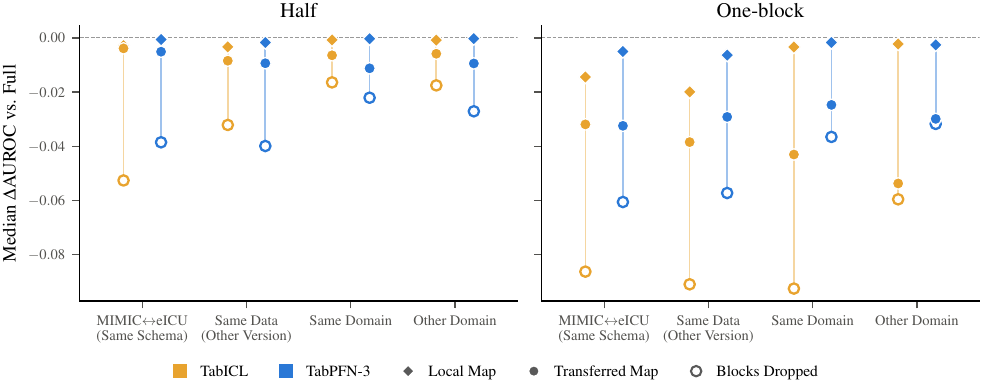}
    \caption{Reusing the map across sites and datasets. Median $\Delta$AUROC vs.\ the full model on the target for the map fitted on the target itself ($\blacklozenge$), a map fitted on another dataset and applied unchanged ($\bullet$, bar: interquartile range over pairs; for the two MIMIC-III$\leftrightarrow$eICU pairs, the range of the two), and dropping the blocks with no map ($\circ$). The closer $\bullet$ lies to $\blacklozenge$, the more of the local map survives transfer. Pairs are averaged over three seeds; all three maps sit at the source's KL-selected window. TabFM, TabDPT and a synthetic source are in \Cref{tab:transfer_grid_full}.}
    \label{fig:transfer}
\end{figure}

Because the shared affine map acts on the schema-free $d$-dimensional residual stream, a transformation fitted on one dataset can theoretically generalize to another. To evaluate this zero-shot transfer, we fit the map on a source dataset's unlabeled activations at the source's own KL-selected window and apply it unchanged to the target at that same window, so no target row is used to place or fit it. In \Cref{fig:transfer} we compare this transferred map against both a locally fitted map at the same window and a naive identity baseline that drops the blocks without a map.

For cohorts sharing an identical feature schema (e.g., the MIMIC-III and eICU intensive-care datasets), the affine map transfers well. At half-depth, a TabICL map fitted on one hospital cohort serves the other with an AUROC penalty of $0.004$, against $0.003$ for a local fit, so it keeps over 90\% of the local map's benefit. However, across 446 disparate dataset pairs (out of 516 total source-target configurations), transfer weakens when the schemas differ, even within one domain (e.g., Heart and Diabetes, both medical but with different features and outcomes). On these pairs, a transferred half-depth map recovers only 60--65\% of a local map's benefit for TabICL and TabPFN-3, about 30\% for TabDPT, and none for TabFM (\Cref{tab:transfer_grid_full}). Domain similarity does not help either: transferring maps from either medical or non-medical sources to MIMIC-III yields identical AUROC penalties for TabICL ($-0.015$ in both cases).  With one block kept, the map transfers weakly, and better within a domain: for TabICL, a map borrowed from the same domain halves the loss of dropping the blocks ($-0.043$ vs.\ $-0.093$), while one from another domain barely improves on it ($-0.054$ vs.\ $-0.060$; \Cref{tab:transfer_grid_full}). At Half, where domain gives no advantage, these results indicate that the skipped transformer blocks execute computations tied to specific features and outcomes, rather than general domain characteristics. A map can therefore be reused across sites or data versions that share a schema; a new schema needs its own local fit, which is fast and label-free.
\Cref{tab:transfer_grid_full} reports the three quantities of \Cref{fig:transfer} for all four models in the transfer grid, with the synthetic source (a Gaussian input with a random two-layer teacher) as a fifth tier. Pairs: MIMIC-III$\leftrightarrow$eICU 2, same data 48, same domain 310 (254 for TabFM), other domain 136, synthetic 20; the other-domain pairs come from 9 sources (the clinical and credit datasets and two physics sets). For all four models a borrowed Half map transfers between versions of the same data and between MIMIC-III and eICU (TabDPT: $-0.013$ from eICU to MIMIC-III, against $-0.002$ local and $-0.063$ for dropping). Across different schemas TabICL and TabPFN-3 keep most of the local map's benefit, while a borrowed map recovers only part of the benefit for TabDPT (same domain: $-0.040$ vs.\ $-0.050$ for dropping the blocks) and none for TabFM ($-0.012$ vs.\ $-0.011$; its Half window costs only $0.01$--$0.02$ even without a map). At One-block, a borrowed map falls far short of a local fit for every model, and the synthetic source breaks TabICL outright ($-0.536$, against $-0.059$ for dropping the blocks). The tiers do not share targets, so they differ in target difficulty as well as in similarity (note the different local and dropped columns). On the 16 targets that receive both same-domain and other-domain sources, the two tiers differ by a median of at most $0.002$ at Half for TabICL and TabPFN-3.

\begin{table}[h]
    \centering
    \caption{Reusing the map across datasets, all models. Median $\Delta$AUROC vs.\ the full model on the target, at the source's KL-selected window, for the map fitted on the target (local), a map fitted on another dataset (transferred) and no map (dropped), by source--target similarity. Pairs are averaged over seeds first: three seeds for TabICL and TabPFN-3, seed~0 for TabFM and TabDPT.}
    \label{tab:transfer_grid_full}
    \small
    \setlength{\tabcolsep}{4pt}
\begin{tabular}{lllrrrrr}
\toprule
Model & Depth & Map & MIMIC$\leftrightarrow$eICU & Same data & Same domain & Other domain & Synthetic \\
\midrule
\multirow{6}{*}{TabICL} & \multirow{3}{*}{Half} & local & $-0.003$ & $-0.003$ & $-0.001$ & $-0.001$ & $-0.001$ \\
 &  & transferred & $-0.004$ & $-0.008$ & $-0.006$ & $-0.006$ & $-0.008$ \\
 &  & dropped & $-0.053$ & $-0.032$ & $-0.016$ & $-0.018$ & $-0.010$ \\
\cmidrule{2-8}
 & \multirow{3}{*}{One-block} & local & $-0.014$ & $-0.020$ & $-0.003$ & $-0.002$ & $-0.002$ \\
 &  & transferred & $-0.032$ & $-0.038$ & $-0.043$ & $-0.054$ & $-0.536$ \\
 &  & dropped & $-0.086$ & $-0.091$ & $-0.093$ & $-0.060$ & $-0.059$ \\
\midrule
\multirow{6}{*}{TabPFN-3} & \multirow{3}{*}{Half} & local & $-0.001$ & $-0.002$ & $0.000$ & $0.000$ & $0.000$ \\
 &  & transferred & $-0.005$ & $-0.009$ & $-0.011$ & $-0.009$ & $-0.007$ \\
 &  & dropped & $-0.039$ & $-0.040$ & $-0.022$ & $-0.027$ & $-0.029$ \\
\cmidrule{2-8}
 & \multirow{3}{*}{One-block} & local & $-0.005$ & $-0.006$ & $-0.002$ & $-0.003$ & $-0.002$ \\
 &  & transferred & $-0.032$ & $-0.029$ & $-0.025$ & $-0.030$ & $-0.034$ \\
 &  & dropped & $-0.061$ & $-0.057$ & $-0.037$ & $-0.032$ & $-0.032$ \\
\midrule
\multirow{6}{*}{TabFM} & \multirow{3}{*}{Half} & local & $-0.007$ & $-0.006$ & $-0.001$ & $-0.002$ & $-0.001$ \\
 &  & transferred & $-0.006$ & $-0.008$ & $-0.012$ & $-0.010$ & $-0.013$ \\
 &  & dropped & $-0.015$ & $-0.021$ & $-0.011$ & $-0.004$ & $-0.054$ \\
\cmidrule{2-8}
 & \multirow{3}{*}{One-block} & local & $-0.009$ & $-0.025$ & $-0.005$ & $-0.006$ & $-0.005$ \\
 &  & transferred & $-0.059$ & $-0.081$ & $-0.110$ & $-0.089$ & $-0.219$ \\
 &  & dropped & $-0.105$ & $-0.171$ & $-0.137$ & $-0.126$ & $-0.124$ \\
\midrule
\multirow{6}{*}{TabDPT} & \multirow{3}{*}{Half} & local & $0.000$ & $-0.003$ & $-0.001$ & $-0.002$ & $-0.002$ \\
 &  & transferred & $-0.025$ & $-0.012$ & $-0.040$ & $-0.021$ & $-0.036$ \\
 &  & dropped & $-0.122$ & $-0.055$ & $-0.050$ & $-0.040$ & $-0.024$ \\
\cmidrule{2-8}
 & \multirow{3}{*}{One-block} & local & $-0.013$ & $-0.017$ & $-0.012$ & $-0.012$ & $-0.012$ \\
 &  & transferred & $-0.118$ & $-0.102$ & $-0.260$ & $-0.215$ & $-0.202$ \\
 &  & dropped & $-0.264$ & $-0.264$ & $-0.433$ & $-0.393$ & $-0.398$ \\
\bottomrule
\end{tabular}
\end{table}

\subsection{Regression} \label{sec:app-regression}

\Cref{tab:regression} gives the per-width medians behind \Cref{fig:regression}, and \Cref{tab:regression_scale} splits the One-block loss into a loss of scale and a loss of ordering.

\begin{table}[h]
    \centering
    \caption{Regression over 41 datasets (MIMIC-III length of stay, 13 TabArena tasks, OpenML-CTR23): median full-model $R^2$ and median $\Delta R^2$ (compressed $-$ full) at the best of widths 3--4 (chosen by calibration $R^2$), at half depth, and with two and one blocks kept; medians over datasets of the mean over 3 seeds.}
    \label{tab:regression}
    \small
\begin{tabular}{lrrrrrr}
\toprule
Model & datasets & Full $R^2$ & best of 3--4 & $L/2$ & $L{-}2$ & $L{-}1$ \\
\midrule
TabPFNv2 (12) & 41 & 0.784 & -0.001 & -0.019 & -0.176 & -0.324 \\
TabPFN-3 (24) & 41 & 0.859 & -0.002 & -0.012 & -0.111 & -0.132 \\
TabICL (12) & 41 & 0.863 & -0.003 & -0.009 & -0.065 & -0.105 \\
TabDPT (16) & 41 & 0.769 & -0.006 & -0.026 & -0.244 & -0.401 \\
TabFM (24) & 41 & 0.873 & -0.000 & -0.001 & -0.016 & -0.022 \\
LimiX-16M (12) & 41 & 0.853 & -0.002 & -0.020 & -0.293 & -0.568 \\
\bottomrule
\end{tabular}
\end{table}

For each regression model, dataset and width we re-ran the calibration-selected window of \Cref{tab:regression} at seed 0 and recorded, besides $R^2$, the Spearman rank correlation of the prediction with the target, the standard deviation of the compressed predictions relative to the full model's, and $R^2$ after an affine recalibration $y \approx a\,\hat y + b$ whose two coefficients are fitted on the compressed predictions of the calibration split (the regression analogue of the classifier's post-hoc scaling; test labels are never used). \Cref{tab:regression_scale} gives the medians. At width 4 nothing moves. With one block kept the predictions are shrunk, the recalibration recovers a third of the loss where the outputs still carry the ordering (TabPFNv2, TabPFN-3, TabICL: on the datasets losing more than $0.05$ $R^2$, the median fraction recovered is $0.33$, $0.36$ and $0.40$) and little for TabDPT ($0.05$), whose ordering collapses, or TabFM ($0.08$), which loses little $R^2$ to begin with, and the rank correlation falls in the same order as the $R^2$ loss. LimiX-16M is the exception: its compressed predictions shrink to $0.22$ of the full model's spread while the rank correlation falls only from $0.92$ to $0.78$, and the recalibration recovers three quarters of its loss (median $\Delta R^2$ $-0.556 \to -0.139$). For the other models the loss is therefore mostly a loss of ordering, with the output scale a secondary and repairable part.

\begin{table}[h]
\centering
\caption{Scale vs.\ ordering decomposition (seed 0, medians over 41 datasets). Recal.\ = $\Delta R^2$ after affine recalibration on the calibration split; $\rho$ = Spearman rank correlation; scale = compressed/full std ratio. Most models lose ordering, not just scale.}
\label{tab:regression_scale}
\small
\setlength{\tabcolsep}{4pt}
\begin{tabular}{llrrrr}
\toprule
Model & Blocks replaced & $\Delta R^2$ & recal.\ & $\rho$ full $\to$ compr.\ & scale \\
\midrule
TabPFNv2 & 4/12 & -.002 & -.000 & .92 $\to$ .91 & 0.99 \\
 & 11/12 & -.333 & -.164 & .92 $\to$ .73 & 0.48 \\
TabPFN-3 & 4/24 & -.001 & -.002 & .93 $\to$ .93 & 0.98 \\
 & 23/24 & -.139 & -.093 & .93 $\to$ .86 & 0.72 \\
TabICL & 4/12 & -.004 & -.005 & .93 $\to$ .92 & 0.98 \\
 & 11/12 & -.095 & -.060 & .93 $\to$ .86 & 0.71 \\
TabDPT & 4/16 & -.007 & -.005 & .91 $\to$ .91 & 0.97 \\
 & 15/16 & -.445 & -.337 & .91 $\to$ .57 & 0.71 \\
TabFM & 4/24 & -.000 & -.000 & .94 $\to$ .93 & 0.99 \\
 & 23/24 & -.023 & -.023 & .94 $\to$ .92 & 0.94 \\
LimiX-16M & 4/12 & -.003 & -.003 & .92 $\to$ .92 & 0.98 \\
 & 11/12 & -.556 & -.139 & .92 $\to$ .78 & 0.22 \\
\bottomrule
\end{tabular}
\end{table}

\subsection{Extended hardware measurements} \label{app:efficiency}

\paragraph{Absorbing the translator into the encoder or the head.}\label{app:folding}
\Cref{sec:method} explains how the map can be absorbed by an adjacent linear layer. Encoder side (window starting at block $0$): in TabPFNv2 each feature token of $\mathbf{H}_0$ is the output of a linear layer, so $\mathbf{H}_0 \mathbf{W}_f + \mathbf{b}_f$ is again a linear function of that layer's input, and the per-feature map is absorbed into a per-feature copy of the encoder's last linear layer. Head side (window ending at block $L$): TabDPT's head reads $\mathbf{H}_L$ through a linear layer, which absorbs the map with its shape unchanged. Every other model has a normalization at the relevant end, which prevents absorption: TabPFN-3, TabICL, TabFM and both Orions normalize the stack output before the head, and the encoders of TabPFN-3, LimiX and Orion-Bix end in a LayerNorm or RMSNorm. Across the full sweep, the selected One-block window touches the absorbing end in 86\% of dataset--seed pairs for TabPFNv2 and 97\% for TabDPT. \Cref{tab:folding} verifies that absorption is exact on the eight clinical datasets at three seeds (fp32 on the GPU): predictions with and without absorption differ by at most $4\times10^{-6}$, and every metric (AUROC, AUPRC, accuracy, balanced accuracy, F1, Brier, ECE) agrees to within $10^{-5}$.

\begin{table}[h]
\centering
\small
    \caption{Absorbing the translator at One-block, eight clinical datasets $\times$ 3 seeds, fp32 on an RTX 5090. Pairs: dataset--seed pairs whose KL-selected window reaches the absorbing end (Heart keeps TabPFNv2's block 0 at two seeds). \emph{Max $|\Delta p|$}: largest difference in predicted probability, with vs.\ without absorption. \emph{Parameters}: translator size before absorption $\to$ extra parameters the absorbed model carries over the one-block model (TabPFNv2: per-feature copies of the encoder's last linear layer; range over datasets).}
    \label{tab:folding}
    \begin{tabular}{llccc}
    \toprule
    Model & Absorbed into & Pairs & Max $|\Delta p|$ & Parameters \\
    \midrule
    TabPFNv2 & encoder & 22/24 & $4\times10^{-6}$ & 0.22--1.45M $\to$ +5--43k \\
    TabDPT & head & 24/24 & $4\times10^{-6}$ & 0.59M $\to$ +0 \\
    \bottomrule
    \end{tabular}
\end{table}

\textbf{CPU deployment.} \Cref{tab:hardware_cpu} repeats \Cref{tab:hardware} on a desktop CPU (Intel i9-14900K, fp32, batch of 512 queries, re-encode regime). There the transformer stack takes most of the forward pass for the lightweight-encoder models, so their end-to-end speed-up approaches the stack's ($9.5$--$12.4\times$ at One-block), while the heavy-encoder models stay at $1.2$--$1.8\times$.

\begin{table}[h]
    \centering
    \small
    \caption{\textbf{Hardware Impact on a CPU.} MIMIC-III on a desktop CPU (Intel i9-14900K, fp32). Blocks params: parameters of the transformer blocks kept, translator included. $\Delta$FLOPs: forward-pass FLOPs removed. End-to-end: batch of 512 queries, context re-encoded on every call.}
    \label{tab:hardware_cpu}
    \small
    \setlength{\tabcolsep}{4pt}
    \resizebox{.8 \textwidth}{!}{%
        \begin{tabular}{l ccc cc cc}
            \toprule
                      & \multicolumn{3}{c}{Blocks params (M)} & \multicolumn{2}{c}{$\Delta$FLOPs} & \multicolumn{2}{c}{End-to-end speed-up}                                                    \\
            \cmidrule(lr){2-4} \cmidrule(lr){5-6} \cmidrule(lr){7-8}
            Model     & Full                                 & Half                              & One-block                               & Half    & One-block & Half        & One-block    \\
            \midrule
            TabPFNv2  & 7.1                                  & 5.0                               & 2.0                                     & $-$50\% & $-$91\%   & 2.2$\times$ & 12.4$\times$ \\
            TabPFN-3  & 51.4                                 & 25.9                              & 2.4                                     & $-$26\% & $-$51\%   & 1.2$\times$ & 1.4$\times$  \\
            TabICL    & 25.7                                 & 13.1                              & 2.4                                     & $-$23\% & $-$42\%   & 1.1$\times$ & 1.3$\times$  \\
            TabDPT    & 75.6                                 & 38.4                              & 5.3                                     & $-$49\% & $-$92\%   & 2.0$\times$ & 11.0$\times$ \\
            TabFM     & 1{,}611                              & 810                               & 71.3                                    & $-$31\% & $-$59\%   & 1.3$\times$ & 1.8$\times$  \\
            Orion-MSP & 25.2                                 & 12.9                              & 2.4                                     & $-$22\% & $-$37\%   & 1.1$\times$ & 1.3$\times$  \\
            Orion-Bix & 25.2                                 & 12.9                              & 2.4                                     & $-$21\% & $-$40\%   & 1.1$\times$ & 1.2$\times$  \\
            LimiX-16M & 15.9                                 & 11.4                               & 4.8                                     & $-$50\% & $-$91\%   & 2.0$\times$ & 9.5$\times$ \\
            \bottomrule
        \end{tabular}}
\end{table}

\begin{table}[h]
\centering
\caption{Stack-only vs.\ end-to-end speed-up on GPU (MIMIC-III, 3{,}000-row context (1{,}500 for TabDPT), one RTX 5090, batch of 512 queries, re-encode regime, median of 10 runs, CUDA synchronised at the stack boundaries). Stack: speed-up of the transformer stack alone; end-to-end: of the whole forward pass. The stack alone still gets close to $L/(\text{blocks kept})$ faster, but on a GPU it is a smaller share of the forward pass than on a CPU, so less of that gain reaches the end-to-end time.}
\label{tab:stack_latency_gpu}
\small
\setlength{\tabcolsep}{4pt}
\resizebox{.9\textwidth}{!}{%
\begin{tabular}{lrrrrrrr}
\toprule
 & & \multicolumn{2}{c}{Width 4} & \multicolumn{2}{c}{Half} & \multicolumn{2}{c}{One-block} \\
\cmidrule(lr){3-4} \cmidrule(lr){5-6} \cmidrule(lr){7-8}
Model & Stack share of time & stack & end-to-end & stack & end-to-end & stack & end-to-end \\
\midrule
TabPFNv2 & 75\% & 1.5$\times$ & 1.38$\times$ & 2.0$\times$ & 1.73$\times$ & 11.2$\times$ & 4.63$\times$ \\
LimiX-16M & 91\% & 1.5$\times$ & 1.43$\times$ & 2.0$\times$ & 1.83$\times$ & 11.5$\times$ & 6.10$\times$ \\
TabDPT & 80\% & 1.2$\times$ & 1.17$\times$ & 1.7$\times$ & 1.52$\times$ & 6.9$\times$ & 3.25$\times$ \\
TabFM & 29\% & 1.2$\times$ & 1.05$\times$ & 2.0$\times$ & 1.17$\times$ & 23.1$\times$ & 1.38$\times$ \\
TabPFN-3 & 19\% & 1.2$\times$ & 1.02$\times$ & 1.9$\times$ & 1.09$\times$ & 19.7$\times$ & 1.23$\times$ \\
TabICL & 10\% & 1.5$\times$ & 1.02$\times$ & 2.0$\times$ & 1.04$\times$ & 10.5$\times$ & 1.09$\times$ \\
Orion-MSP & 13\% & 1.5$\times$ & 1.05$\times$ & 1.9$\times$ & 1.07$\times$ & 9.6$\times$ & 1.15$\times$ \\
Orion-Bix & 26\% & 1.5$\times$ & 1.09$\times$ & 2.0$\times$ & 1.15$\times$ & 11.1$\times$ & 1.31$\times$ \\
\bottomrule
\end{tabular}}
\end{table}

\subsection{nanoTabPFN training setup}
\label{app:nano_setup}

\begin{figure}[h]
    \centering
    \includegraphics[width=.9\textwidth]{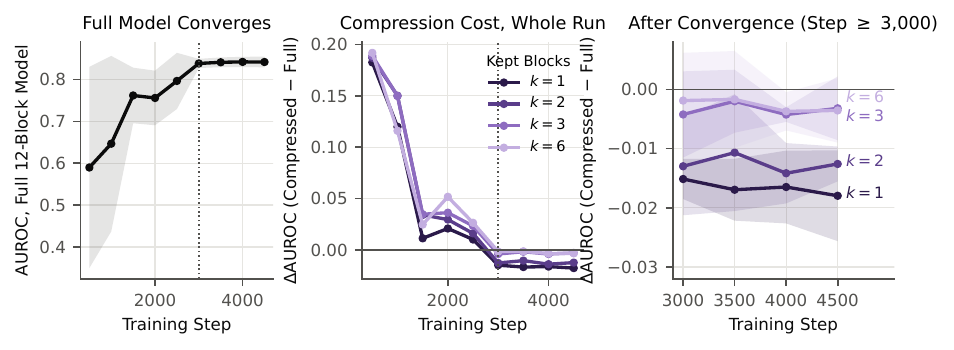}
    \caption{nanoTabPFN training dynamics: loss curves, evaluation AUROC and learning-rate restarts across depths $L \in \{1, 2, 3, 6, 12\}$ (three seeds each). Deeper models ($L \geq 6$) collapse to uniform predictions at the default learning rate and require an automatic restart at a quarter of the rate.}
    \label{fig:nano_dynamics}
\end{figure}

\Cref{fig:nano_dynamics} shows the training dynamics across depths. \Cref{tab:train_shallow} lists the values plotted in \Cref{fig:train_shallow}.

\begin{table}[!ht]
    \centering
    \caption{\textbf{Pre-training shallow vs.\ compressing deep}, values of \Cref{fig:train_shallow}. nanoTabPFN test AUROC, averaged over four clinical datasets; mean $\pm$ std over three seeds.}
    \label{tab:train_shallow}
    \small
    \begin{tabularx}{\textwidth}{l *{5}{>{\centering\arraybackslash}X}}
        \toprule
        Kept blocks $k$              & 1    & 2    & 3    & 6    & 12   \\
        \midrule
        Trained at depth $k$         & $.850_{\pm.009}$ & $.902_{\pm.008}$ & $.894_{\pm.005}$ & $.902_{\pm.001}$ & $.903_{\pm.003}$ \\
        Trained, compressed topology & $.852_{\pm.008}$ & $.888_{\pm.008}$ & $.900_{\pm.004}$ & $.900_{\pm.002}$ & ---  \\
        Compressed $12 \to k$        & $.854_{\pm.012}$ & $.891_{\pm.001}$ & $.903_{\pm.005}$ & $.902_{\pm.007}$ & ---  \\
        Compressed $12 \to k$, prior-fitted map & $.428_{\pm.158}$ & $.550_{\pm.116}$ & $.832_{\pm.091}$ & $.756_{\pm.202}$ & ---  \\
        \bottomrule
    \end{tabularx}
\end{table}

We use the nanoTabPFN reference code unchanged except for two additions. Every model has embedding size 96, 4 attention heads and an MLP hidden size of 192, from 132k parameters at $L{=}1$ to 1.37M at $L{=}12$. It is trained on a public set of pre-generated prior tasks (300k synthetic binary tasks, 150 rows $\times$ 5 features) for 5{,}000 steps at batch size 32 with schedule-free AdamW (no weight decay, gradient-norm clipping at 1), and the weights are taken at the optimizer's averaged point. The tasks are read in order, so every variant and seed sees the same sequence of prior batches and the seed changes only the initialisation. The first addition skips prior batches containing NaNs: the set contains five corrupt tasks, and the upstream NaN check is disabled, so a single such batch would ruin the weights. The second is a collapse check that uses the training loss alone, so no evaluation data is used during training. A run that falls into the uniform-prediction basin stays at the chance loss of the binary prior ($\ln 2 \approx 0.693$) for the whole run. If the mean training loss over the last 500 steps is not below $\ln 2 - 0.02$, the run is restarted from the same seed at a quarter of the learning rate. Starting from $4 \times 10^{-3}$, all depths $L \leq 3$ train at that rate at every seed, while $L \in \{6, 12\}$ collapse at every seed and train at $10^{-3}$. No run needed a second restart. As a control, we also retrained the $L \in \{1,2,3\}$ models at $10^{-3}$ (the learning rate required for the 12-layer model). Their mean AUROC shifts by $-0.002$, $-0.019$ and $+0.007$, respectively; the $k=2$ shift comes from one of the three seeds. The compressed models still match or slightly exceed these shallow models ($0.854$ vs.\ $0.849$ at $k=1$; $0.891$ vs.\ $0.883$ at $k=2$; $0.903$ vs.\ $0.901$ at $k=3$), so the learning-rate difference does not favor compression. The affine variant (``trained, compressed topology'') inserts one $96 \times 96$ linear layer, initialised to the identity, after the first $\lceil L/2 \rceil$ blocks of an $L$-block model and trains it jointly. Its learning rates follow the same rule. For evaluation, each dataset is split per seed into a 100-row support, a calibration set of at most 150 rows and a test set of at most 1{,}000 rows. The five features are the top-5 by F-score on the support. The 12-to-$k$ compression uses the paper's ridge recipe, with the window selected on the calibration set.

\subsection{Context extrapolation in nanoTabPFN} \label{app:nano_context}

We repeat the comparison on the two large clinical cohorts, MIMIC-III and eICU. At a support of 100 rows the picture is unchanged. We then raise the support to 1{,}000 rows, about $7\times$ the 150-row tasks seen in pre-training, so the model must extrapolate in context length. Depth does not help here either: with $k \geq 2$ kept blocks the compressed models stay within $0.008$ of the full 12-layer model on both cohorts ($0.651$--$0.660$ vs.\ $0.656$ on eICU, $0.674$--$0.681$ vs.\ $0.682$ on MIMIC-III). Only the one-block model loses ($0.660$ on MIMIC-III, $0.630$ on eICU). Shallow models trained from scratch do at least as well as the full model ($0.660$--$0.667$ on eICU and $0.694$--$0.701$ on MIMIC-III for $k \geq 2$), and at $k=2$ they lead the compressed models by $0.016$ (eICU) and $0.027$ (MIMIC-III).

\end{document}